%% file: main.tex
\documentclass[letterpaper]{article} % DO NOT CHANGE THIS
\usepackage{aaai25}  % DO NOT CHANGE THIS
\usepackage{times}  % DO NOT CHANGE THIS
\usepackage{helvet}  % DO NOT CHANGE THIS
\usepackage{courier}  % DO NOT CHANGE THIS
\usepackage[hyphens]{url}  % DO NOT CHANGE THIS
\usepackage{graphicx} % DO NOT CHANGE THIS
\usepackage{natbib}  % DO NOT CHANGE THIS AND DO NOT ADD ANY OPTIONS TO IT
\usepackage{caption} % DO NOT CHANGE THIS AND DO NOT ADD ANY OPTIONS TO IT
\usepackage{algorithm}
\usepackage{algorithmic}
\usepackage{multirow,tabularx}
\usepackage{booktabs}
\usepackage{amsmath}
\usepackage{amssymb}
\usepackage{subcaption}
\usepackage{color}
\usepackage{dsfont}
\usepackage{tcolorbox}
\tcbuselibrary{breakable,listings}
\newtcblisting{longprompt}{
    breakable,
    colback=gray!10,
    colframe=gray!30,
    listing only,
    listing options={
        basicstyle=\ttfamily\small,
        breaklines=true,
        breakatwhitespace=false,
        postbreak=\mbox{}, % No indentation for wrapped lines
        columns=flexible,
        keepspaces=true,
        numbers=none, % Explicitly disable line numbers
        showlines=false, % Hide empty lines at the end
        breakindent=0pt % Ensure no indentation after line breaks
    },
    left=3pt,
    right=3pt,
    top=3pt,
    bottom=3pt,
    boxsep=2pt,
    width=\columnwidth
}
\usepackage{newfloat}
\usepackage{listings}
\DeclareCaptionStyle{ruled}{labelfont=normalfont,labelsep=colon,strut=off} % DO NOT CHANGE THIS
\lstnewenvironment{prompt}[1][]
{
    \lstset{
        basicstyle=\ttfamily\small,
        breaklines=true,
        breakatwhitespace=false,
        postbreak=\mbox{\textcolor{red}{$\hookrightarrow$}\space},
        columns=flexible,
        keepspaces=true,
        frame=single,
        framesep=3pt,
        rulecolor=\color{gray!30},
        backgroundcolor=\color{gray!10},
        numbers=none,
        numberstyle=\tiny\color{gray},
        numbersep=5pt,
        xleftmargin=0pt,
        xrightmargin=0pt,
        aboveskip=5pt,
        belowskip=5pt,
        #1
    }
}
{}
\floatstyle{ruled}
\newfloat{listing}{tb}{lst}{}
\floatname{listing}{Listing}
\title{Assessing Modality Bias in Video Question Answering Benchmarks with Multimodal Large Language Models}

\author{
    Jean Park\textsuperscript{\rm 1},
    Kuk Jin Jang\textsuperscript{\rm 1},
    Basam Alasaly\textsuperscript{\rm 2},
    Sriharsha Mopidevi\textsuperscript{\rm 2},
    Andrew Zolensky\textsuperscript{\rm 1},\\
    Eric Eaton\textsuperscript{\rm 1},
    Insup Lee\textsuperscript{\rm 1},
    Kevin Johnson\textsuperscript{\rm 1,2}
}
\affiliations{
    \textsuperscript{\rm 1 }Department of Computer and Information Science, University of Pennsylvania \\
    \textsuperscript{\rm 2} Perelman School of Medicine, University of Pennsylvania\\
    Philadelphia, PA\\
    \{hlpark, jangkj, zolensky, eeaton, lee\}@seas.upenn.edu,\\ \{basam.alasaly, sriharsha.mopidevi, kevin.johnson1\}@pennmedicine.upenn.edu
}

\usepackage{bibentry}
\begin{document}

\maketitle

\begin{abstract}
Multimodal large language models (MLLMs) can simultaneously process visual, textual, and auditory data, capturing insights that complement human analysis. 
However, existing video question-answering (VidQA) benchmarks and datasets often exhibit a bias toward a single modality, despite the goal of requiring advanced reasoning skills that integrate diverse modalities to answer the queries.

In this work, we introduce the modality importance score (MIS) to identify such bias. It is designed to assess which modality embeds the necessary information to answer the question. 
Additionally, we propose an innovative method using state-of-the-art MLLMs to estimate the modality importance, which can serve as a proxy for human judgments of modality perception.
With this MIS, we demonstrate the presence of unimodal bias and the scarcity of genuinely multimodal questions in existing datasets. 
We further validate the modality importance score with multiple ablation studies to evaluate the performance of MLLMs on permuted feature sets. 
Our results indicate that current models do not effectively integrate information due to modality imbalance in existing datasets. 
Our proposed MLLM-derived MIS can guide the curation of modality-balanced datasets that advance multimodal learning and enhance MLLMs' capabilities to understand and utilize synergistic relations across modalities.

\end{abstract}

\input{introduction}
\input{related_work}

\input{method}
\input{evaluation}
\input{limitation}
\input{conclusion}
\bibliography{aaai25}
\clearpage
\appendix
\input{appendix}

\end{document}

%% file: introduction.tex
\section{Introduction}

%figure about complementary question and modality-agnostic questions 
\begin{figure*}[!htb]
    \centering
        \begin{subfigure}[b]{0.7\textwidth}
        \centering
        \includegraphics[width=\textwidth]{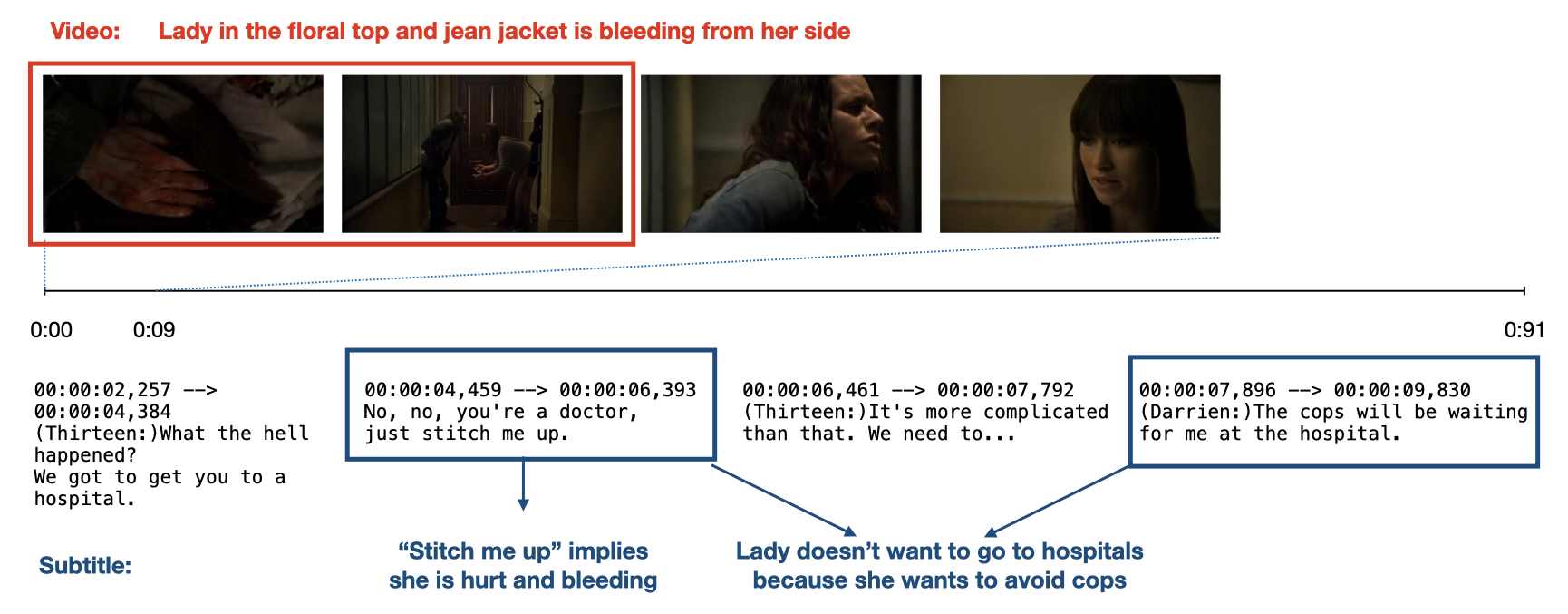}
        \caption{Video and subtitle (TVQA)}
        \label{fig:motiv_video}
    \end{subfigure}
    \begin{subfigure}[b]{0.26\textwidth}
        \centering
        \includegraphics[width=\textwidth]{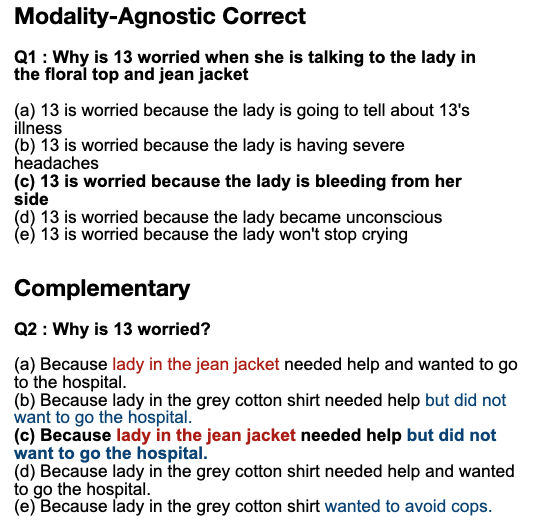}
        \caption{Example questions}
        \label{fig:motiv_questions}
    \end{subfigure}
    \caption{Example of a video clip with multimodal questions demonstrating different modality importance. $Q_1$ is answerable using either subtitle or video information, while $Q_2$ requires integrating information from both modalities. (Sec. \ref{sec:category})}
    \label{fig:motivating_example}
\end{figure*}

\begin{figure}[!htb]    
    
\end{figure}

In recent years, trends in AI development have leaned towards multimodal models, particularly multimodal large language models (MLLMs), as many complex problems necessitate the integration of diverse modalities to achieve more accurate and comprehensive reasoning.

Video question answering (VidQA) stands out as a particularly challenging task, requiring the integration of various modalities along with complex spatial and temporal reasoning \cite{xiao2021next}.
As such, this task serves as a vital benchmark for assessing the vision-language understanding capabilities of AI systems.

In recent years, several VidQA benchmarks have been developed to train and evaluate the capabilities of MLLMs in these areas \cite{yu2019activitynet, gupta2022vquad}. 
However, a fundamental question remains: Are these models genuinely integrating information from various sources, or are they simply leveraging biases inherent in the datasets?
Our observations suggest that many existing benchmarks are limited in their ability to assess this integration.
The questions often tend to be biased toward a single modality, or \emph{modality bias}, lacking the complexity that would require genuine multimodal integration.
For instance, the video question $Q_1$ depicted in Fig. \ref{fig:motiv_questions} can be answered using only the video alone or the subtitles alone. 
Although having redundant information across modalities may be beneficial for learning cross-modal relationship, it doesn't fully represent the complexity of real-world multimodal reasoning tasks. 

As illustrated in $Q_2$ from Fig. \ref{fig:motiv_questions}, some multimodal questions require integrating distinct pieces of information from the text (not wanting to go to the hospital) and from the video (material of clothing) to accurately deduce the answer.
Unfortunately, such questions that demand genuine integration of multiple modalities are notably scarce in current datasets.

To address these limitations, we need a method that quantitatively assesses modality bias in questions. 
To this end, we introduce a novel {\bf modality importance score (MIS)}, which evaluates the extent to which each modality contributes to answering a given question. 
Using this score, we perform a comprehensive assessment of modality bias in existing VidQA benchmarks. Our analysis reveals significant limitations in current datasets and highlights the need for more balanced and challenging multimodal questions. 

Our main contributions are as follows:

\begin{itemize}

\item We propose a novel modality importance score (MIS) 

and a method that leverages multimodal large language models (MLLMs) to estimate the MIS. 
We show that this approach could serve as a proxy for human judgements of modality perception. 
\item Using the proposed modality importance score, we demonstrate the existence of a unimodal bias and the scarcity of truly multimodal questions in current multimodal datasets. 
\item We evaluate several state-of-the-art multimodal models on questions with permuted features for modalities with low importance scores. The results reveal that current multimodal models do not optimally combine information from different sources due to modality imbalance in existing multimodal datasets.
\end{itemize}

By addressing these limitations in VidQA benchmarks, our work aims to advance the field of multimodal AI, pushing towards models that can genuinely integrate information across modalities to perform complex reasoning tasks more effectively.

%% file: related_work.tex
\section{Related Work}\label{section:related_work}

\subsection{Video Question Answering}
Video question answering (VidQA) is a well-explored field in AI, presenting the challenge of integrating multimodal input from videos, understanding temporal and causal relations, and selecting the correct answer \cite{lei2019tvqa+}. 
Many recent VidQA models are pretrained on large datasets using contrastive learning objectives~\cite{kim2021self}, masked language modeling~\cite{fu2021violet}, and other techniques to learn joint representations and improve spatio-temporal understanding~\cite{zhao2017video,jiang2020divide}.
These models are subsequently fine-tuned on downstream tasks, such as open-ended or multiple choice video-question answering~\cite{wang2023all}, video-text retrieval~\cite{luo2020univl}, and video captioning~\cite{fu2023empirical}.

In this study, we focus on four approaches that have been developed to utilize both subtitle and video information for video question answering. 
\textbf{Merlot Reserve}~\cite{zellers2022merlot} is pretrained to predict either the correct text or audio snippet hidden by a MASK token, given uniformly sampled images from a video. 
Its architecture includes pretrained encoders for each modality input and a joint encoder trained with a contrastive spanning objective. 
\textbf{FrozenBiLM}~\cite{yang2022frozenbilm} employs a frozen bidirectional language model trained on web-scale multimodal data.
\textbf{Llama-VQA}~\cite{ko-etal-2023-large} builds upon the Llama model, incorporating additional learnable parameters through the Flipped-VQA framework. 
This approach leverages the LLM's prior knowledge of temporal and causal reasoning. 
\textbf{MiniGPT4-Video}~\cite{ataallah2024minigpt4} is an open-source multimodal large language model designed for video-language tasks. 
Its training process involves pretraining using either Llama2 or Mistral on video-text pairs consisting of frame sequences and subtitles appended to a predefined prompt. 
In addition, other VidQA approaches utilize captions and videos, such as VindLu or MMFT-BERT, and MSAN~\cite{cheng2023vindlu,khan2020mmft, kim2020msan}.
Additional tasks and approaches outside the scope of this study can be found in a survey by \citet{zhong2022video}.

While these models show improved performance by integrating language and video inputs for video understanding, a critical issue remains: they are trained on datasets that have questions with modality bias.
This bias raises the question of whether these models can leverage both modalities for each question and context and whether they are biased in their ability to leverage either modality as appropriate.
Our research examines whether current models can effectively identify and use the most relevant modality, even with irrelevant information.
Our findings reveal limitations in their ability to perform this task optimally.

\subsection{VidQA Datasets and Benchmarks}
Several notable datasets and benchmarks have been proposed for multiple-choice VidQA approaches.

\subsubsection{TVQA} The TVQA dataset~\cite{lei2018tvqa} comprises over 150K question-answer pairs derived from 21,793 clips across six TV shows. These clips average 76 seconds, with each question providing a localized timestamp indicating where the answer can be found within the clip. 

In TVQA's test-public set, human accuracy varied across different modality combinations: 61.96\% for video-only, 73.03\% for subtitles-only, and 89.41\% for both. 
While the authors interpret this result as evidence for the necessity of both visual and textual understanding, we propose an alternative perspective. 
We hypothesize that many questions in the dataset contain redundant information across both video and subtitle sources rather than requiring the integration of information from distinct sources.
Furthermore, we believe this result insufficiently captures how questions depend on different modalities or their combinations.
\subsubsection{LifeQA}
The LifeQA dataset~\cite{castro-etal-2020-lifeqa} comprises 2.3K questions derived from 275 real-life YouTube videos. These videos were recorded by individuals in uncontrolled environments, capturing meaningful visual and linguistic interactions. The human performance on this dataset varied significantly: when given only video, participants achieved 48.5\% accuracy; with audio alone, accuracy rose to 63.4\%; and with all modalities combined, accuracy peaked at 90.6\%. 
Interestingly, these results contradict the authors' Venn diagram (\citet{castro-etal-2020-lifeqa}, Fig. 3) categorization of LifeQA questions by answer type.
Their categorization suggests that over 60\% of questions are visual-based, while only 29\% are speech-based, with the remaining questions (10\%) requiring both modalities. This distribution seems at odds with the observed human performance across different modality combinations.
We argue this discrepancy suggests that the authors' categorization of answer types may have been based on the perceived nature of the question rather than actual modality dependency.
This method may be less accurate, as some questions labeled as ``Visual'' like ``Where are they located?'' might also be answered based on dialogue or background sounds.

\subsubsection{AVQA}
The AVQA (Audio-Visual Question Answering) dataset~\cite{yang2022avqa}, derived from the VGG Sound dataset~\cite{chen2020vggsound}, contains over 57K question-answer pairs derived from 57K real-life videos focusing on object-generated sounds rather than human speech. 
It was designed to require information from both audio and visual modalities for most questions, to ensure that relying on just one modality would be insufficient or ambiguous for an accurate answer. 
However, the annotators who designed the questions also categorized the question types.
Similar to LifeQA, this approach could introduce bias, as annotators might focus on the perceived modality requirements rather than objectively assessing whether relevant information is present in each modality.

\subsection{Modality Contribution in Multimodal Tasks}
The concept of quantifying modality contributions in multimodal tasks was explored in perceptual score paper~\cite{gat2021perceptual}. They introduced a ``perceptual score'' to measure a model's reliance on specific input modalities or subsets. Their method involved removing the influence of a modality M from the set of all modalities and measuring the resulting change in accuracy.

Others, such as  \citet{yang2024quantifyingenhancingmultimodalrobustness}, revealed that multimodal models often prefer certain modalities, leading to less robust performance when a modality is missing or perturbed. Their research showed that models tend to rely on one specific modality even when trained on multiple modalities, demonstrating vulnerability to unimodal attacks. To address this issue, they introduced Certifiable Robust Multi-modal Training, a method designed to mitigate the influence of the model's modality preference and regulate essential components to improve its robustness. 

While such works aim to analyze models' bias towards specific modalities and suggest solutions for reliable and robust performance, our work focuses on quantifying the modality contribution in the dataset, specifically in multiple-choice VidQA datasets. We identify modality bias in these datasets and provide a more fine-grained categorization of question types. This approach aims to guide the development of more balanced datasets, a crucial first step toward enabling multimodal models to utilize modalities effectively.

%% file: method.tex
\section{Method} \label{section:method}
\subsection{Modality Importance Score}
\subsubsection{Intuition.}
Understanding the contribution of each modality is crucial in multi-modal question-answering tasks.
Our goal is to distinguish between questions answerable by a single modality, those with redundant signals from multiple modalities, and those requiring integration of modalities.

Consider the scenario in Figure~\ref{fig:motivating_example}, with two input modalities: video and subtitles (audio in the form of text). 
Three input combinations are possible:
video alone, subtitle alone, and video + subtitle. 
The importance of a modality, such as video, can be quantified by estimating the increase in accuracy when video is present in the input combination (video, video+subtitle) relative to when it is not (subtitle).  

In Figure~\ref{fig:motiv_questions}, the question $Q_1$ is an example where accuracy does not increase when the video is added. 
From the phrase, ``stitch me up'' in the subtitle, one can reasonably infer that the lady is likely bleeding. 
The video confirms this fact by displaying a bleeding lady, but adds redundant signals rather than providing essential new details.
In contrast, question $Q_2$,
 exemplifies a multimodal question that cannot be answered correctly with a single modality. 
 When considering only the video, two answer choices, (a) and (c), become confusing as both mention the correct visual detail ``lady in the jean jacket''.
 Similarly, with only subtitles, three plausible answer choices are given (b), (c), and (e).
 The question requires integrating information from both modalities for an accurate response. 
We formalize this intuition by defining the modality importance score.

\subsubsection{Definition.}

Given an input question $q_i$, its corresponding ground truth label $y_i$, and a set of source modalities $M = \{m_1, m_2, ..., m_k\}$, we denote combinations of modalities in $M$ as the power set of $M$ excluding the $\emptyset$, $\mathcal{P}(M) \setminus \emptyset$. 
We first define the performance measurement function as: 
\begin{equation}\label{eqn:perf}
\mathit{perf}(q_i \mid M') = \frac{\sum\limits_{S \subseteq M'} \mathds{1}[\text{A}^{i}_S = {y_i}]}{|M'|} \enspace ,
\end{equation}
where  $M'$ is a subset of modalities defined as $M' \subseteq \mathcal{P}(M) \setminus \emptyset$, and $|M'|$ is the cardinality. 
$\mathds{1}[\text{A}^{i}_S = {y_i}]$ is the response accuracy function we use to measure the performance in VidQA tasks defined as, 
\begin{equation}\label{eqn:indicator_func}
\mathds{1}[\text{A}^{i}_S = {y_i}]  =
\begin{cases}
    1 & \text{if } \text{A}^{i}_S = {y_i} \\
    0 & \text{if } \text{A}^{i}_S \neq {y_i} \enspace.
\end{cases}
\end{equation} 
This is an indicator function that returns 1 if the answer for question $q_i$, $A^i_S$, obtained using a subset of modalities $S$ matches the ground truth, and 0 otherwise.
While our current performance measurement function $\mathit{perf}(q_i \mid M')$ considers only response accuracy, it can be generalized to incorporate other performance metrics. 

Finally, the \textbf{Modality Importance Score (MIS)} for a single modality $m_j$ and question $q_i$, is defined as: 
\begin{equation}\label{eqn:mi}
\begin{split}
    \mathrm{MIS}^{i}_{m_j} = \mathit{perf}(q_i \mid M_{j}^{+}) - \mathit{perf}(q_i \mid  M_{j}^{-}) \enspace ,
\end{split}
\end{equation}
where $M_{j}^{+} = \{S \subseteq \mathcal{P}(M) \setminus \{\emptyset, \{m_j\}\} : m_j \in S\}$ are all the non-empty subsets of modalities that must include $m_j$ excluding the singleton set containing only $m_j$ and $M_{j}^{-} = \{S' \subseteq \mathcal{P}(M) \setminus \{\emptyset, \{m_j\}\} : m_j \not\in S' \}$ are all non-empty subsets of modalities that exclude $m_j$. 

This formulation captures two key aspects of modality importance. The $\mathit{perf}(q_i \mid M_{j}^{+})$ calculates the average accuracy across all subsets of modalities that include $m_j$ and at least one other element from the set of modalities in $M$, capturing how well $m_j$ contributes in combination with other modalities. 
The $\mathit{perf}(q_i \mid M_{j}^{-})$ computes the average accuracy across all subsets that exclude $m_j$. 
The difference measures the overall impact of including $m_j$ versus excluding it. 
Note that our intention is to compute the modality importance score for a single modality $m_j$ and not a set of multiple modalities; however, it is trivial to expand the definitions of $ M+$ and $ M-$ to include or exclude combinations of multiple modalities.

\begin{table}[!ht]
\centering
% \resizebox{0.6\columnwidth}{!}{
\begin{tabular}{ccc|cc}
\toprule
\multicolumn{3}{c|}{Response Accuracy} & \multirow{2}{*}{$\mathrm{MIS}_{\text{Vid}}$} & \multirow{2}{*}{$\mathrm{MIS}_{\text{Sub}}$}\\
$\text{Vid}$ & $\text{Sub}$ & $\text{Vid} + \text{Sub}$&  \\ \midrule
0  & 0 & 0 &  0 & 0 \\ \hline
0  & 1 & 0 &  -1 & 0 \\ \hline
1  & 0 & 0 &  0 & -1 \\ \hline
1  & 1 & 0 &  -1 & -1 \\ \hline
0  & 0 & 1 &  1 & 1 \\ \hline
0  & 1 & 1 &  0 & 1 \\ \hline
1  & 0 & 1 &  1 & 0 \\ \hline
1  & 1 & 1 &  0 & 0 \\ \bottomrule                  
\end{tabular} 
% }
\caption{Modality Importance Score for Two Individual Modalities : Video (Vid), Subtitle (Sub) }
\label{tab:mi_score_table}
\end{table} 

Table~\ref{tab:mi_score_table} 
illustrates modality importance scores for response accuracies of three modality combinations. The scores can be interpreted as follows: Positive MIS indicate that the modality embeds a signal contributing to the answer beyond other modalities. Negative MIS suggest that the modality adds interference of conflicting information, potentially masking another modality's contribution. An MIS of 0 implies that the modality doesn't contribute additional information beyond other modalities.

Note that the MIS reflects a modality's relative contribution compared to others, not its absolute ability to answer a question. For instance, if the subtitle alone can answer a question, the video's MIS may be 0, indicating no additional contribution, and vice versa. In such cases, the modality subset with both modalities might have MIS of 0, reflecting their redundancy rather than their inability to answer the question.

\subsubsection{MLLM-derived Modality Importance Score}
To estimate the modality importance for questions in dataset $D$, we can leverage the capabilities of MLLMs for scalability purposes. 
This approach is applicable to datasets with $|M|\geq 2$ modalities.

For each combination, we prompt the MLLM to select the most plausible answer choice given the provided input combination. 
We compare the model's response accuracy across different input combinations and quantify the relative importance of each modality according to \eqref{eqn:mi}.

This approach provides insights into the distribution of critical information across modalities in multimodal question-answering tasks.
Previous approaches~\cite{gat2021perceptual}, 
used random permutation to simulate the removal of a modality's influence due to the complexity of altering trained models. 
Our approach does not require permutation as MLLMs allow for more direct manipulation of input modalities. 
Although our MIS metric can quantify each individual modality's contribution when more than two modalities are present, current MLLMs typically support only images and text. 
Hence, for this study, we compute modality importance providing three distinct input combinations to the MLLM: subtitle only, video only, and both modalities together.

\subsection{Categorizing Question Types with MIS}
\label{sec:category}
\subsubsection{Unimodal-bias questions}
Using the MIS, we can identify unimodal-biased questions. 
If $\mathrm{MIS}^i_{m_k} \leq 0 \leq \mathrm{MIS}^i_{m_j} , \enspace \mathrm{MIS}^i_{m_k} \neq \mathrm{MIS}^i_{m_j}$  
$\forall m_k \in M$ where $m_k \neq m_j$, we classify question $q_i$ as $m_j$-biased. 
Such questions can be answered using only $m_j$, but cannot be answered correctly using any other single modality $m_k$.
For instance, with video and subtitle modalities, video-biased questions can manifest in two ways. 
First, correct answers might be obtained whenever the video modality is included, but using only subtitles leads to incorrect answers due to their irrelevance. 
Alternatively, the video alone might yield correct answers, but combining video and subtitles could result in incorrect answers. 
In this latter case, the MIS for subtitles becomes negative, indicating interference.

\subsubsection{Modality-agnostic vs Complementary questions} 
In addition to identifying unimodal-biased questions, we use MIS to provide a more fine-grained categorization of questions. 
This categorization helps our understanding of multimodal questions and the relationships between different modalities in answering them.

\paragraph{Modality-agnostic Question} As shown in Table~\ref{tab:mi_score_table} rows 1 and 8, there are cases where the same MIS is obtained regardless of which the subset of modalities, correct or incorrect.
We define these as \textbf{modality-agnostic questions}, where $\forall m_j \in M,  \mathrm{MIS}_{m_j}=0$. 
We further divide modality-agnostic questions into two subcategories: 
\begin{itemize}
    \item Modality-agnostic correct questions: \\
    $\forall S \subseteq \mathcal{P}(M)$, $\mathds{1}[\text{A}^{i}_S = {y_i}] = 1$
    \item Modality-agnostic incorrect questions: \\
    $\forall S \subseteq \mathcal{P}(M)$, $\mathds{1}[\text{A}^{i}_S = {y_i}] = 0$ 
\end{itemize}

\paragraph{Complementary Questions} As illustrated in row 5 of Table~\ref{tab:mi_score_table}, there exist questions where no single modality can strongly determine the answer and signals from multiple modalities can be combined to determine the correct answer. 
We define these questions as \textbf{complementary questions}, where $\forall m_j \in M, \mathrm{MIS}_{m_j} > 0$. 
In this case, all modalities contribute to answering the question correctly when combined with other modalities. 

Note that in the case of only two modalities, complementary questions cannot be answered correctly unless both modalities are utilized.
For scenarios with more than two modalities, complementary questions may involve varying contributions from each modality. 

%% file: evaluation.tex
\section{Evaluation}\label{section:evaluation}

\subsection{Experimental Setup and Overview} 

\subsubsection{Estimating modality importance score}
For our experiments, we utilized GPT-4 Turbo~\cite{openai2024gpt4technicalreport}, one of the top-performing MLLMs that supports both image and text inputs. 
We prompted the model to select the correct answer by providing the question, answer choices, and the corresponding modality combination under evaluation.
Specific constraints and image extraction were applied to account for GPT-4 Turbo's token limitations and allow longer video clips.
Detailed information about our prompts and process can be found in Appendix~\ref{section:appendix}.

\subsubsection{Datasets}
We evaluated three VidQA datasets, each containing both video and subtitle/audio components. For TVQA~\cite{lei2018tvqa} and LifeQA~\cite{castro-etal-2020-lifeqa}, we use transcripts/subtitles provided by the dataset. AVQA~\cite{yang2022avqa} does not provide transcripts, but we use the audio labels from VGG Sound dataset~\cite{chen2020vggsound} as the subtitle. 
Due to the large number of questions, we limited evaluation to the validation or test sets. For TVQA and AVQA, we uniformly sampled 
1,019 and 796 questions, respectively, representing approximately 6-10\% of the total questions. For LifeQA, we evaluated the entire test set of 372 questions. 

\subsubsection{VidQA Models}
Our study evaluates four multimodal VidQA models, listed in Table~\ref{tab:Mm_permute_evaluation}, capable of processing both visual and textual (audio captions or subtitle) inputs to answer multiple-choice questions. 
We use the MLLM-derived MIS to identify unimodal-biased questions. 
Our feature permutation experiments show how effectively these models integrate and utilize information across different modalities.

\subsection{Human Study Validation of MLLM-derived Modality Importance}\label{subsection:human_eval}
To assess human perception of modality importance, we employed a split-group methodology involving four participants, each evaluating 197 TVQA questions.
The detailed methodology is in Appendix~\ref{section:appendix}, along with Figure~\ref{fig:human_study} depicting the study and Table~\ref{tab:accuracy_confidence} showing accuracy distributions across confidence levels.
Our study aimed to validate the alignment between MLLM-derived MIS and human perception of modality importance. 
The evaluation yielded a substantial inter-annotator agreement (Fleiss' kappa: 0.76) for questions answered with both modalities, with an average accuracy of 87.8\%.

\begin{figure}[!htb]
    \centering
    \includegraphics[width=0.35\textwidth]{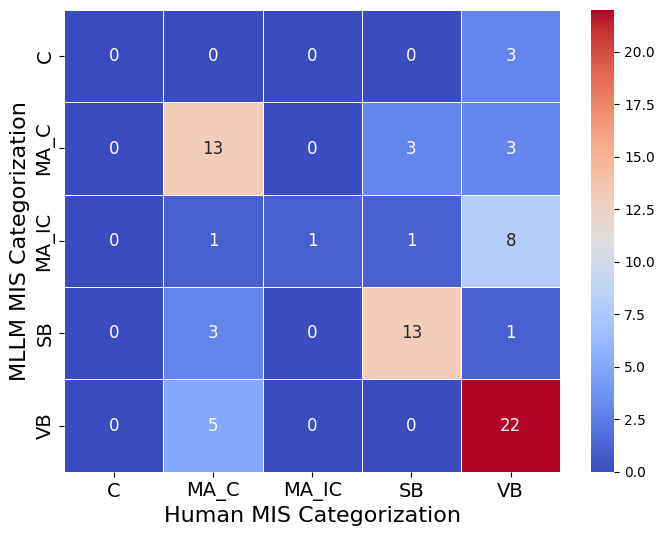}
    
    \caption{Question categorization based on human study vs MLLM-derived MIS}
    \label{fig:heatmap_mi_strict}
\end{figure}

For our analysis, we focused on questions that showed unanimous agreement per modality. 
For these questions annotators were either all correct or all incorrect. As shown in Fig. ~\ref{fig:heatmap_mi_strict}, this method revealed a strong alignment between human perception-based and MLLM-derived categorizations for three types of questions: modality-agnostic correct, subtitle-biased, and video-biased.
This suggests that when human annotators are clearly in agreement, their judgments closely match the MLLM's assessments.

Under this categorization based on human scores, we were unable to identify any complementary questions from the evaluated subset of questions. 
This observation suggests that questions whose answer relies on information from both modalities might indeed be scarce in the multimodal VidQA dataset. 
This finding highlights a potential limitation in current multimodal datasets. 

\subsection{Evaluation of Modality Bias in VidQA Datasets}\label{subsection:vidqa_eval}
In this section, we analyze the distribution of question types based on MLLM-derived MIS. 

\begin{table*}[!ht]
\centering
\begin{tabular}{l|cccccc|c}
\toprule
\multirow{2}{*}{} & \multicolumn{6}{c|}{\# of Q per Question Types} & \multirow{2}{*}{Total \# of Q} \\ 
                  & $\text{SB}$      & $\text{VB}$      & $\text{C}$      & ${\text{MA}}_{\text{C}}$  & ${\text{MA}}_{\text{IC}}$ &   $\text{None}$   & \\ \midrule
TVQA              & 224 (22.0\%) & 345 (33.9\%) & 21 (2.1\%) & 357 (35.1\%) & 71 (7.0\%) & 1 (0.1\%)& 1019 \\ \hline
LifeQA            & 74 (19.9\%) & 125 (33.6\%) & 9 (2.4\%) & 135 (36.3\%) & 29 (7.8\%) & 0 (0.0\%) & 372 \\ \hline
AVQA              & 39 (4.9\%) & 93 (11.7\%) & 5 (0.6\%) & 625 (78.5\%) & 32 (4.0\%) & 4 (0.5\%) & 796 \\\bottomrule                  
\end{tabular} 

\caption{Distribution of Question Types based on MIS Across Different Datasets : Question (Q), Subtitle-biased (SB), Video-biased (VB), Complementary (C), Modality-agnostic Correct ($\text{MA}_\text{C}$), Modality-agnostic Incorrect ($\text{MA}_\text{IC}$)}
\label{tab:MI_distribution}
\end{table*} 

\subsubsection{TVQA} 
The results, reported in Table~\ref{tab:MI_distribution}, support our assumption that many questions in TVQA would be modality-agnostic correct. 
About 35\% of the questions were classified as modality-agnostic correct, while only 2\% were identified as complementary, requiring information from both modalities. 
We had 7\% of questions that were modality-agnostic incorrect. 
As shown in Figure~\ref{fig:heatmap_mi_strict}, GPT has limited visual understanding compared to humans, as 8 out of 11 modality-agnostic incorrect questions were actually video-biased. 
While the subtitle does not provide relevant information for these questions, GPT fails to extract or comprehend details from the sequence of images. 
Consequently, the model consistently incorrect regardless of input modality. 
Overall, the results show a potential discrepancy between the dataset's intended multimodal nature and the actual distribution of question types.

\begin{figure}[!htb]
    \centering
    \begin{subfigure}[b]{0.23\textwidth}
        \centering
        \includegraphics[width=\textwidth]{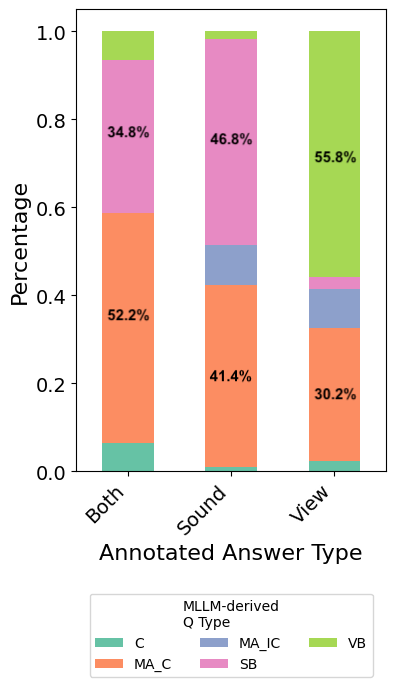}
        \caption{LifeQA question types}
        \label{fig:lifeqa_qtype}
    \end{subfigure}
    \hfill
    \begin{subfigure}[b]{0.23\textwidth}
        \centering
        \includegraphics[width=\textwidth]{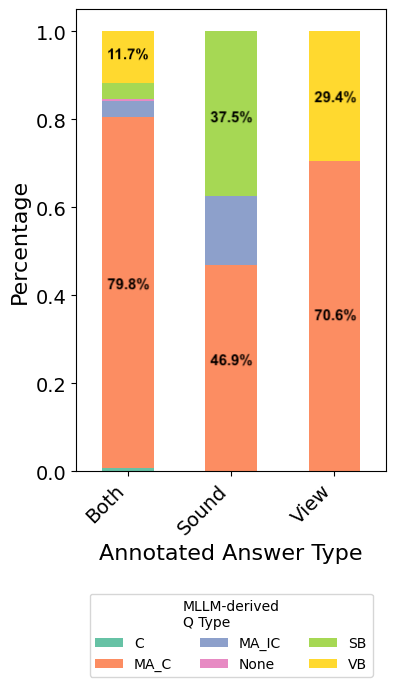}
        \caption{AVQA question types}
        \label{fig:avqa_qtype}
    \end{subfigure}
    \caption{Proportion of MIS based Question types per Annotated Answer Type}
    \label{fig:qtype_proportion}
\end{figure}

\subsubsection{LifeQA}

The distribution of question types based on our MIS categorization shown in Table~\ref{tab:MI_distribution} revealed that modality-agnostic correct questions formed the largest category, accounting for approximately 36\% of the dataset.
Video-biased questions followed closely, comprising 33\% of the dataset, and subtitle-biased questions accounted for 19.9\%. 
Less than 10\% of questions were modality-agnostic incorrect for ``Sound'' and ``View'' types.
For ``View'' types, we found out that GPT-4's had limitations in identifying image details.
For ``Sound'' types, errors were primarily due to insufficient information in the provided automated captions.
The low percentage of complementary questions (2\%) indicates that most questions in the LifeQA dataset can be answered using a single modality or are modality-agnostic. 

 Figure~\ref{fig:lifeqa_qtype} compares our MI-based categorization with the annotated answer types.
- For ``Sound'' answer types, 46.8\% were classified as subtitle-biased, aligning with the annotated type. 
However, a significant 41\% were categorized as modality-agnostic. 
This suggests that many questions annotated as language-dependent can actually be answered with all modalities. 
Similarly, for ``View’' answer type questions, while the majority were video-biased, a significant number were modality-agnostic correct. 
These observations indicate that our categorization generally aligns with human-annotated answer types.
Moreover, the significant proportion of modality-agnostic correct questions in both ``Sound'' and ``View'' answer types suggests that many questions may not be single modality-dependent. See Appendix~\ref{section:appendix} for examples. 

\begin{figure}[!htb]
    \centering
    \includegraphics[width=0.47\textwidth]{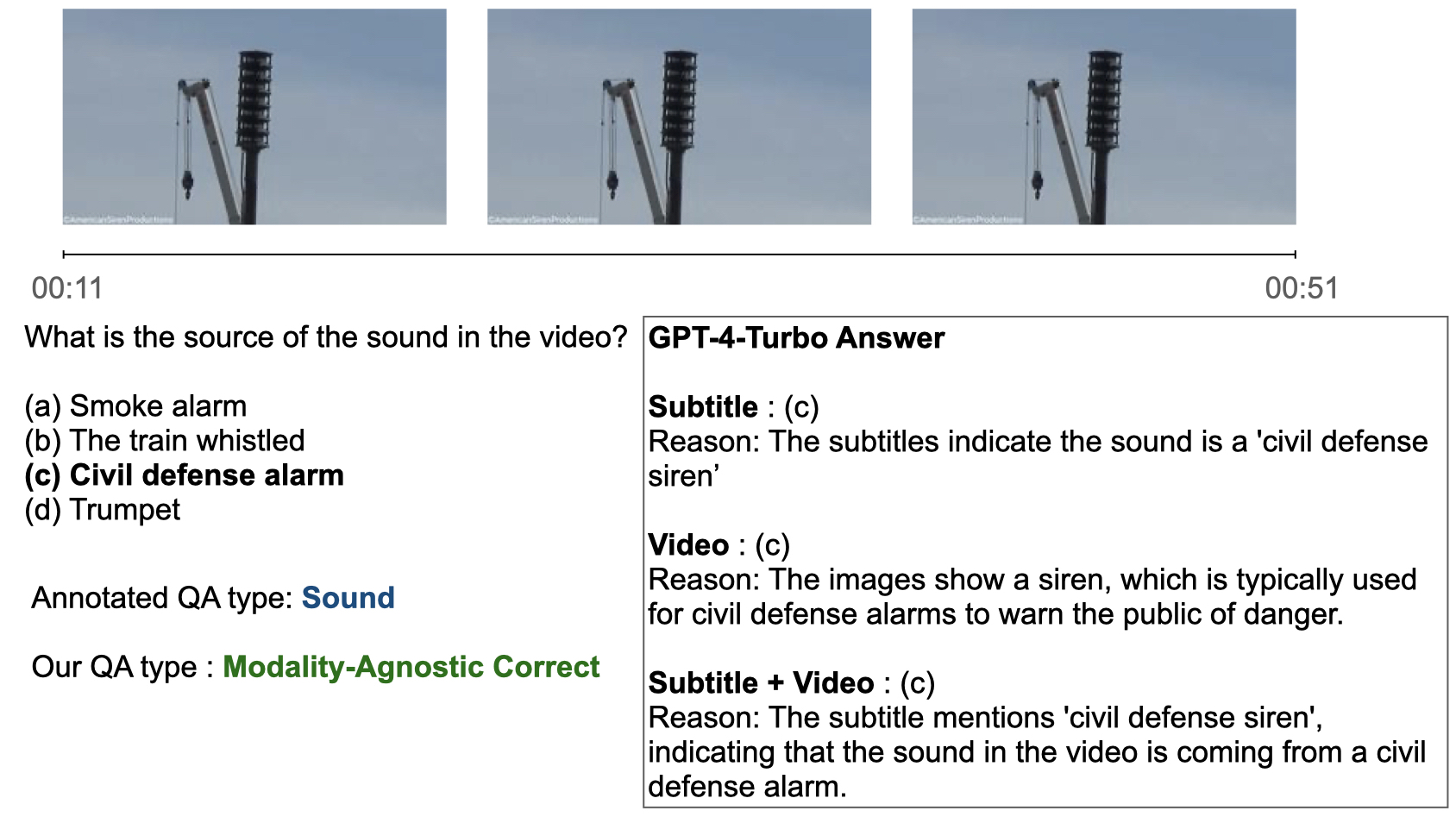}
    \caption{Example from AVQA where annotated answer type is different from our categorization. For this video, the subtitle is ``civil defense siren''.
    }
    \label{fig:avqa_example}
\end{figure}

\subsubsection{AVQA}
Table~\ref{tab:MI_distribution} depicts our analysis of AVQA. 
Our analysis found that the distribution of question types contradicts the dataset's original design intention of requiring both modalities to answer accurately.
78.5\% of 796 questions were modality-agnostic correct questions. 
This implies that many questions in this dataset are answerable using any single modality, as shown in Figure~\ref{fig:avqa_example}.
Only a small fraction of questions, approximately 0.6\%, were complementary. 
Additional examples can be found in Appendix~\ref{section:appendix}. 

Figure~\ref{fig:avqa_qtype} reveals interesting patterns similar to LifeQA. 
Based on our categorization, questions annotated with the ``Sound'' answer consisted of 37.5\% subtitle-biased questions and no video-biased questions. 
Similarly, the "Video" answer type questions showed a high number of video-biased questions (29.4\%) and no subtitle-biased questions.

\subsubsection{Summary}
In summary, our study demonstrates that the MLLM-derived MIS and question categorization align well with human perception of modality relevance. This approach shows that many seemingly single-modality questions are modality-agnostic correct, indicating the presence of redundant signals across modalities. Although our sampling method prevented us from definitively proving dataset-wide unimodal bias, our approach shows significant potential in identifying such biases and highlighting the scarcity of truly multimodal questions requiring sophisticated information integration from multiple modalities. 

\begin{table*}[!htbp]
\centering
\begin{tabular}{l|ccc|ccc}
\toprule
\multirow{2}{*}{} & \multicolumn{3}{c|}{Subtitle-biased} & \multicolumn{3}{c}{Video-biased}\\ 
        & Orig.  & SP ($\Delta$) & VP ($\Delta$) & Orig. & SP ($\Delta$) & VP ($\Delta$) \\ \midrule
Merlot R\text{*} & 91.5 $\pm$ 0.0 & 32.2  $\pm$ 3.8 (-59.3) & 87.4  $\pm$ 1.9 (-4.1) & 71.9 $\pm$ 0.0 & 72.0  $\pm$ 1.5 (+0.1) & 43.2  $\pm$ 5.0 (-28.7)\\ \hline
FrozenBiLM & 95.5 $\pm$ 0.0 & 31.3 $\pm$ 4.3 (-64.2) & 96.3  $\pm$ 0.3 (+0.8) & 75.4 $\pm$ 0.0 & 73.4 $\pm$ 2.7 (-1.9) & 41.5 $\pm$ 4.4  (-33.9) \\ \hline
Llama-VQA & 95.1  $\pm$ 0.0 & 37.3 $\pm$ 1.8 (-57.8 ) & 94.3  $\pm$ 0.0 (-0.8) & 56.9 $\pm$ 0.0 & 56.1 $\pm$ 0.3 (-0.8) & 47.5  $\pm$ 1.5 (-9.4) \\\hline
MiniGPT4\text{*} & 61.4 $\pm$ 0.2 & 35.9 $\pm$ 3.6 (-25.5) & 58.7 $\pm$ 3.5  (-2.8) & 42.4 $\pm$ 0.8  & 40.9 $\pm$ 2.0 (-1.5)& 38.6 $\pm$ 3.2 (-3.9) \\ \midrule
Average & 85.9  $\pm$ 0.0 & 34.2 $\pm$ 3.4 (-51.7) & 84.2 $\pm$ 1.5 (-1.7)& 61.6 $\pm$ 0.2 & 60.6 $\pm$ 1.6 (-1.0) & 42.7 $\pm$ 3.0 (-19.0)  \\ \bottomrule 
\end{tabular} 
\caption{Accuracy (\%) comparison after feature permutation with five random seeds (Orig: Original, SP: Subtitle permuted, VP: Video permuted, Merlot R\text{*}: Merlot Reserve, MiniGPT4\text{*} : MiniGPT4-Video). All models except for MiniGPT4-Video were fine-tuned on TVQA dataset.}
\label{tab:Mm_permute_evaluation}
\end{table*}

\subsection{Multimodal Model Evaluation}

Using the MIS, we partition the TVQA questions into those that exhibit bias towards subtitles or video content to assess the multimodal capability of models.
We perform feature permutation experiments to evaluate how well the models focus on information relevant to each question type. 

The results presented in Table~\ref{tab:Mm_permute_evaluation} demonstrate the effectiveness of MIS in capturing unimodal bias across different models. 
We observe that permuting features with low MIS leads to a significantly smaller decrease in accuracy than permuting features with high MIS. 
For instance, with the subtitle-biased question, {\it ``Why did Marshall think they should have their marriage waiting period waived?''} 
First, we permute the less important video features by providing the correct subtitle with the wrong images from a different TV show. 
Then, we permuted the more important feature by providing the wrong subtitle with the correct images. 
If our MIS effectively categorizes the questions, we would expect the model to perform well in the former case but fail in the latter. 
This expectation aligned with our results, as the average decrease in accuracy between low-MIS and high-MIS feature permutations was 32\%, considering both subtitle and video-biased questions. 

Our evaluation reveals several key insights. 
First, the significant decrease in accuracy between low and high-importance feature permutations confirms that our modality importance score effectively identifies unimodal-biased questions.
Second, models generally show degraded performance on video-biased questions than subtitle-biased ones.
This difference suggests a limitation in understanding visually relevant features across the evaluated models. 
This may be due to the prevalence of subtitle-biased and modality-agnostic questions in the original TVQA datasets. 
Although we were unable to determine the total number of unimodal-biased questions in the TVQA dataset, we can infer from human performance on the TVQA test set. 
In the original TVQA results, human accuracy with subtitles exceeded that with video by 11\%, encompassing both subtitle-biased and modality-agnostic questions. 
Consequently, we hypothesize that models were trained to focus more on subtitles than video. 
This is also supported by our observation that permuting video in video-biased questions resulted in a lower accuracy decrease than permuting subtitles in subtitle-biased questions.
Our additional experiments in Appendix \ref{subsection:MAC_C_analysis} further validate this hypothesis, showing that even when both modalities contain informative signals, models predominantly rely on textual information.
Lastly, when we permuted features with low importance scores, all models showed decreased accuracy except FrozenBiLM with the subtitle modality and Merlot Reserve with video modality. 
This observation indicates that most models struggle to optimally combine information from different modalities, even when one modality is deemed less important for a given question.
These findings highlight the challenges in multimodal learning and the need for improved strategies in integrating information across modalities. 

%% file: limitation.tex
\section{Discussion and Limitations}
The main limitation of our study is the use of a single MLLM, though a small-scale verification with multiple MLLMs (Appendix~\ref{subsection:cross_MLLM_validation}) supports our claim that most questions are modality-agnostic, with few complementary. Our approach is also constrained by the MLLM’s visual processing limitations, likely affecting the categorization of some video-biased questions. Future studies should factor MLLM performance into MIS computation for more robust bias assessment and a more comprehensive evaluation of modality importance in multimodal datasets.

%% file: conclusion.tex
\section{Conclusion}
Our findings reveal a significant challenge in the field of multimodal AI: current Video Question Answering datasets may not be optimally enabling multimodal reasoning. 
Our novel method for assessing the relative importance of different modalities, the MLLM-derived MIS, shows that across three VidQA benchmarks, a substantial 89.8\% to 94.8\% of questions can be answered using a single modality or are modality-agnostic. 
Only 0.6\% to 2\% require genuine multimodal integration. 
Our analysis shows that our MLLM-derived MIS correlates with the human perception of modality importance, suggesting its potential for guiding the scalable curation of more balanced datasets. 
Based on these findings, we propose two future directions: creating new benchmarks that include complementary questions to properly train and evaluate multimodal reasoning, and developing models with dynamic modality integration mechanisms~\cite{kim2020msan} to effectively combine information across modalities.

%% file: appendix.tex
\onecolumn
\section{Appendix} \label{section:appendix}
We present the following items in the appendix:
\begin{itemize}
    \item Experimental Setup
    \item Evaluation Prompts
    \item Human Study Validation of MLLM-derived Modality Importance Regarding Confidence Score
    \item Example Questions from Evaluated Datasets
    \item Configuration for Evaluated Mutlimodal Models
     \item Modality Dependence Analysis through Complementary and Modality Agnostic Questions
    \item Cross-MLLM Validation of Question Type Classification
    \item Answer Choice Order Bias
\end{itemize}

% Upon acceptance of this paper, we will make our code and data processing scripts publicly available.

\subsection{Experimental Setup}\label{subsection:appendix_experimental_setup}
We set GPT's parameters to top-p=0 and seed=123, although the GPT API doesn't guarantee deterministic behavior across runs. 

To accommodate GPT-4 Turbo's token limitations, we implemented the following constraints: 
For subtitles, we did not impose any limitations as all were less than the maximum tokens. 
For video-only inputs, we limited the number of images to 10.
For combined video and subtitle inputs, we reduced the image limit to 8.

Given that many clips in our dataset exceed one minute in duration, we adopted a systematic approach to image extraction:
we sampled frames at 1 Hz, starting from the provided localized timestamp.
For clips exceeding the image number limit, we parsed them into multiple segments and prompted the model to analyze each segment separately.
If the correct answer was identified in any segment, we considered the overall response correct.

\subsection{Evaluation Prompts}
\begin{longprompt}
You are tasked with answering a question with five multiple-choice options for a clip. For each clip, you will be given a question and five answer choices, along with the subtitles from the video.

Select the most likely answer from the given choices based solely on the information provided in the [Input Modality]. Do not make assumptions or rely on external knowledge. If the [Input Modality] do not contain enough information to confidently answer the question, choose the answer that is most plausible given the limited context.

In addition to selecting the most likely answer, specify the [Input Modality's Content Segment] where the relevant information for the correct answer can be found. Also, state the reason you chose the answer. The reason should be no longer than two sentences. If you made a random guess because you were not able to select any plausible answer, then put 'None' in the [Input Modality's Content Segment] but keep the random answer and state the reason as "Could not find answer, I selected random answer.".  

For each video clip, format your output as follows: 
"{ "Question ID 1": {
            "Q":"How did ~?", 
            "Answer Candidates" : {
                "a": "", "b": "", "c": "", "d": "", "e": "" 
            },
            "Answer": "b",
            "[Input Modality's Content Segment]": [],
            "Reason": "The answer ..."
        },
    "Question ID 2": {}
}"
\end{longprompt}
We utilize the above prompt for evaluation, adapting it to various input combinations: subtitles only, video only, or both subtitles and video. 
The phrase ``Input Modality's Content Segment'' in the prompt refers to different elements depending on the given modality. 
For subtitles, it indicates timestamp ranges; for video, it denotes image numbers; and when both are present, it includes both timestamp ranges and image numbers. 
This approach allows us to assess GPT-4's ability to identify relevant information from subtitles and/or video when selecting the correct answer.

For each prompt, we append the question, answer choices, and corresponding input modalities. 
When subtitles are involved, we extract the relevant subtitle text that overlaps with the localized timestamp from TVQA. 
To optimize API request costs, we group five questions, their answer choices, and associated subtitles into a single prompt for subtitle-only evaluations. 
For video-based evaluations, whether video-only or video with subtitles, we adopt a different approach. In these cases, we include only one question and its answer choices per prompt, accompanied by the corresponding video frames. 
When evaluating both video and subtitles together, we follow the same structure as video-only prompts but additionally incorporate the relevant subtitle text.

Given this prompt, GPT-4-Turbo successfully outputted the correct JSON format. However, although we gave clear instructions in the prompt that the model should choose answer from the input choices, the model did not consistently follow these instructions. In cases where it couldn't find the answer, it frequently outputted ``None'' or ``selected random answer'' for the ``answer'' field. We regarded these responses as ``incorrect''.

\subsection{Human Study Validation of MLLM-derived Modality Importance Regarding Confidence Score}

\begin{figure}[!htb]
    \centering
    \includegraphics[width=0.7\textwidth]{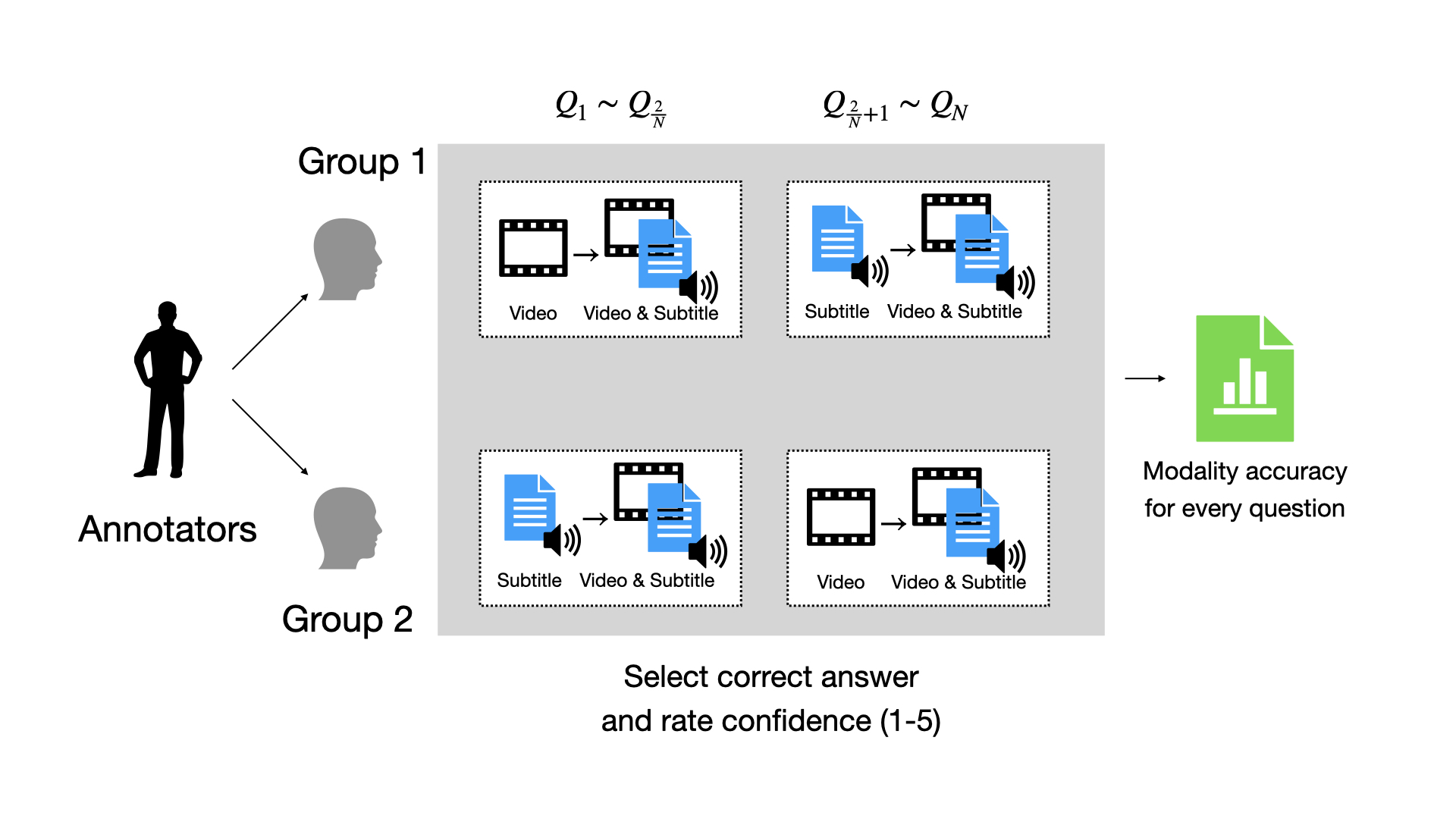}
    \caption{Human modality perception study}
    \label{fig:human_study}
\end{figure}

As shown in the Figure~\ref{fig:human_study}, our human study involved four participants divided into two groups, assessing a total of 197 questions from TVQA. These questions were sampled from the 1,019 questions evaluated (see Section~\ref{subsection:vidqa_eval}), ensuring representation across all categories. 
Each group was presented with the same set of questions but different single-modality inputs initially, followed by combined modality input. To account for the confidence level of responses, we asked participants to rate their confidence for each answer (1-5).

While our evaluation process yielded substantial inter-annotator agreement, 0.76, and the average accuracy of 87.8\% with both modalities, we identified a significant variance between annotators' confidence. This is observed in accuracy scores in the low confidence group in Table~\ref{tab:accuracy_confidence}.

%group confidence score and show accuracy & agreement
\begin{table}[!ht]
\centering
\begin{tabular}{l|ccc}
\toprule
& \multicolumn{3}{c}{Accuracy (\%)}  \\ 
& max & min & mean  \\ \midrule 
    all  & 92.9 & 79.2 &  87.8\\ 
    high confidence  & 95.2 & 87.1 &  92.5\\ 
    low confidence   &  60.0 & 11.1 & 38.6\\ \bottomrule                  
\end{tabular} 

\caption{Human accuracy per confidence score using both modalities (high confidence : confidence $>$ 3, low confidence : confidence $\leq$ 3)}
\label{tab:accuracy_confidence}
\end{table} 

We then calculated a weighted accuracy score by multiplying each individual's accuracy with their confidence score normalized by maximum confidence score of 5. Then we sum these products, and divide by the number of people who used that modality. This weighted accuracy was then rounded to either 0 or 1, considering a modality to contain a strong signal for answering the question if the average accuracy exceeded 0.5.

\begin{figure}[!htb]
    \centering
    \includegraphics[width=0.5\textwidth]{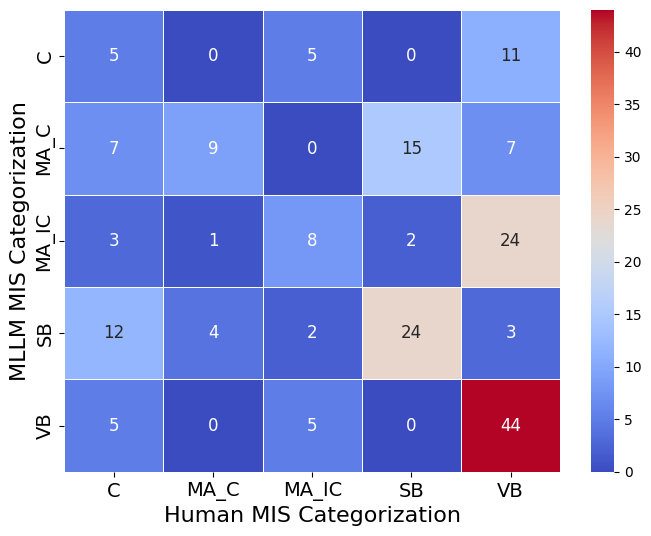}
    \caption{Heatmap of Question Categorization Based on Human Study vs MLLM-derived MIS Score}
    \label{fig:heatmap_mi_estimate}
\end{figure}

Using these rounded response accuracies, we computed the modality importance score across all modality combinations. Our findings indicate that human perception of unimodal bias in questions aligns similarly with MLLM-based assessments as shown in Figure ~\ref{fig:heatmap_mi_estimate}. The correlation between the model's categorization and human judgment demonstrated a Cohen's kappa score of approximately 0.3.

While this score indicates moderate agreement, several factors contribute to the observed variance:
\begin{enumerate}
\item Limited Sample Size: Our study involved only four participants due to resource constraints. This small sample size may not fully capture the diversity of human perceptions and could contribute to the variance in results.

\item Dataset Imperfections: Misaligned subtitles and incorrect speaker information in the TVQA dataset may have led to discrepancies between human and MLLM interpretations.

\item Background Knowledge Disparities: Despite instructions to avoid using external knowledge, MLLMs demonstrated their use of background information to infer scenes and characters' behaviors. This reveals that MLLMs leveraged their extensive knowledge about TV show characters, potentially enabling them to answer questions that some human annotators found challenging due to limited familiarity with specific characters or plot elements.

\end{enumerate}

Despite these factors, the moderate agreement between human and MLLM-based assessments is encouraging, especially considering the study's limitations. It suggests that our computational approach captures significant aspects of human-like understanding of modality relevance in complex video question-answering tasks.

\subsection{Example Questions from Evaluated Dataset}
Our approach identified several interesting examples of modality-agnostic correct responses and complementary questions in both the LifeQA and AVQA datasets. These examples provide valuable insights into the nature of multimodal questions and the performance of MLLM like GPT-4 in video question answering tasks.

\begin{figure}[!htb]
    \centering
    \begin{subfigure}[b]{0.6\textwidth}
        \centering
        \includegraphics[width=\textwidth]{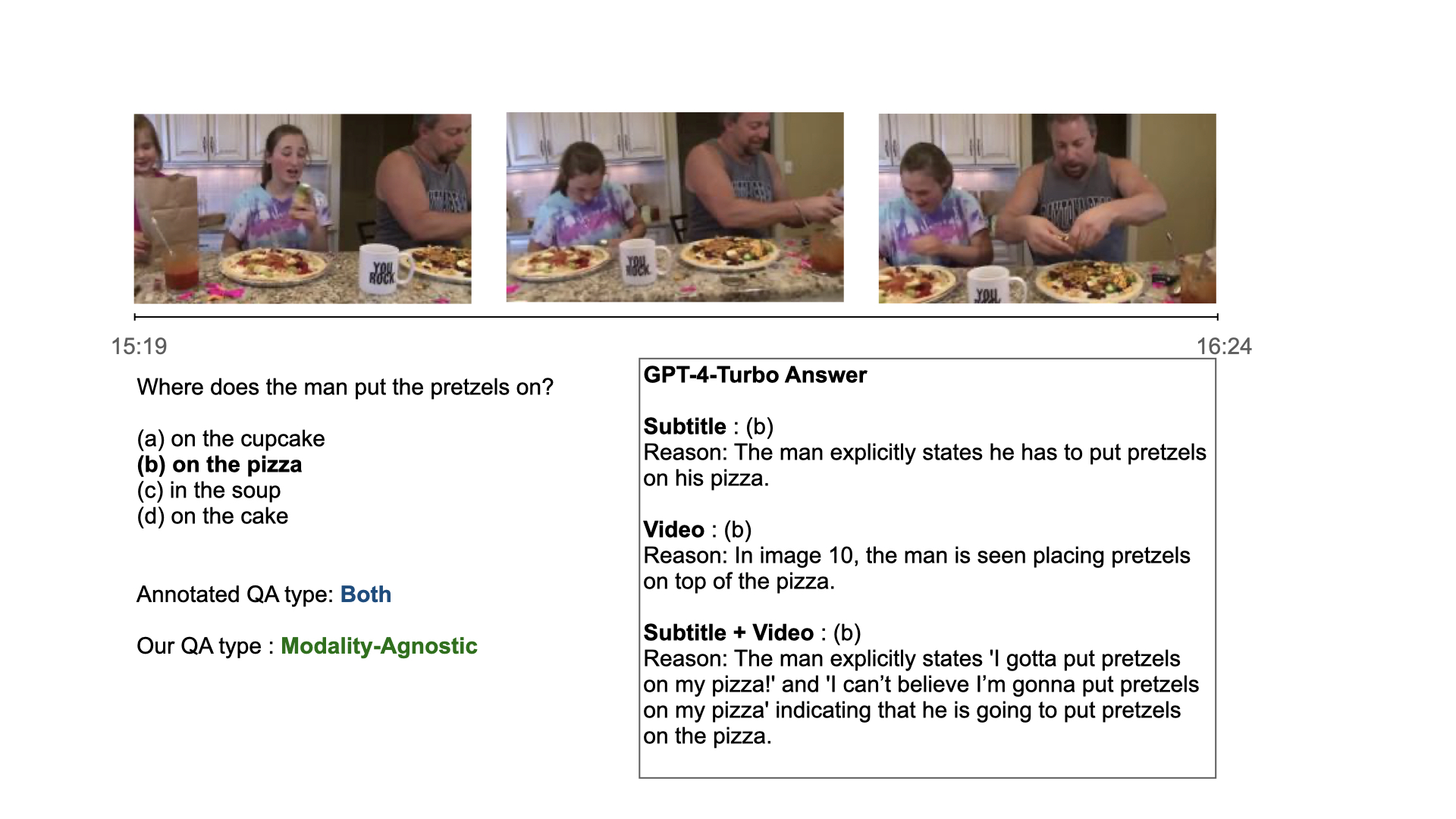}
        \caption{Annotated as ``Both''. Subtitle mentions ``[dad] haha pretzels I gotta put pretzels on my pizza!''.}
        \label{fig:lifeqa_b_ma}
    \end{subfigure}
    \hfill
    \begin{subfigure}[b]{0.6\textwidth}
        \centering
        \includegraphics[width=\textwidth]{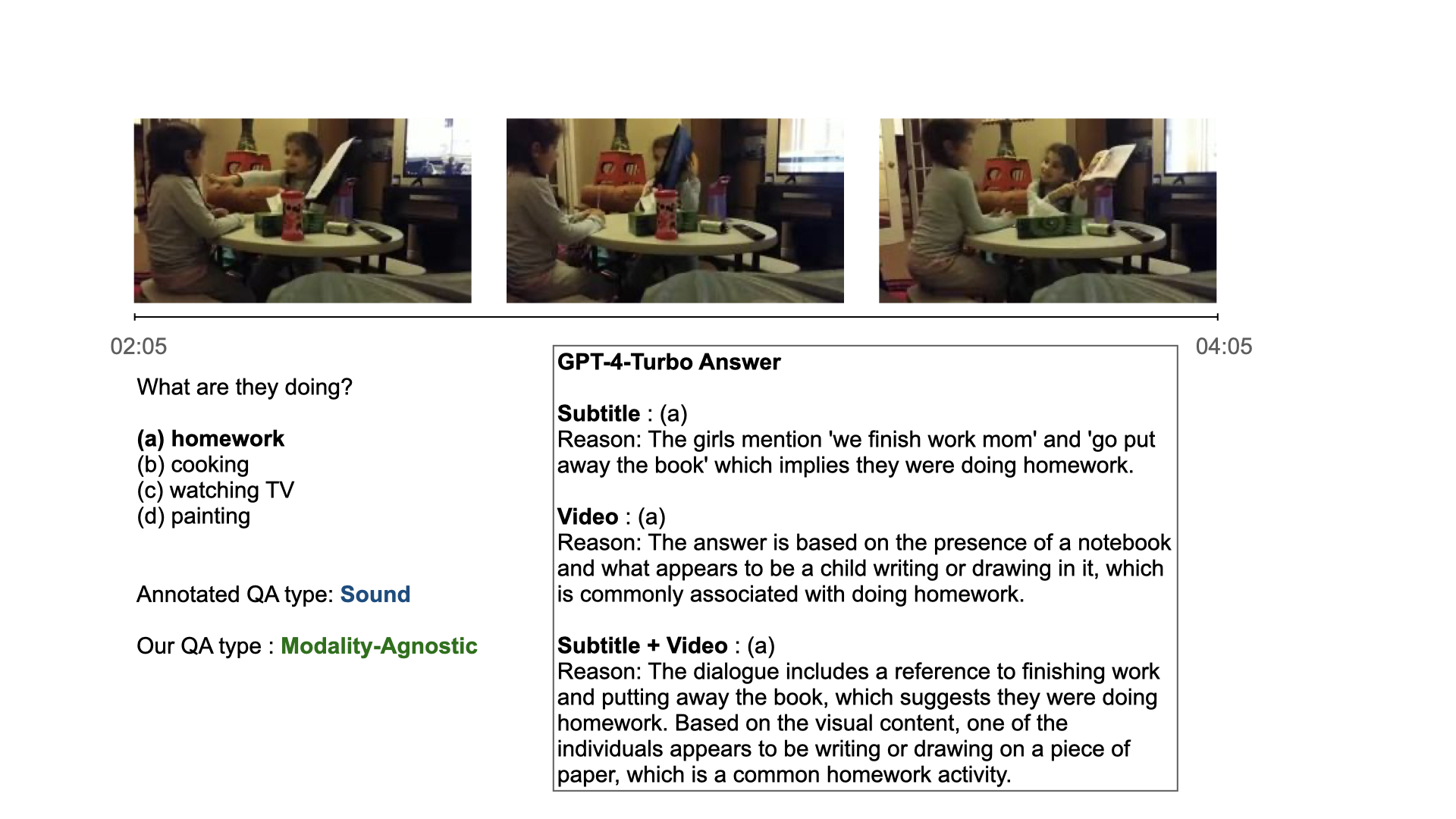}
        \caption{Annotated as ``Sound''. Subtitle mentions ``[girl1] hahaha all right [girls1]  we finish work mom [inaudible] [woman] go put away the book''.}
        \label{fig:lifeqa_s_ma}
    \end{subfigure}
    \hfill
    \begin{subfigure}[b]{0.6\textwidth}
        \centering
        \includegraphics[width=\textwidth]{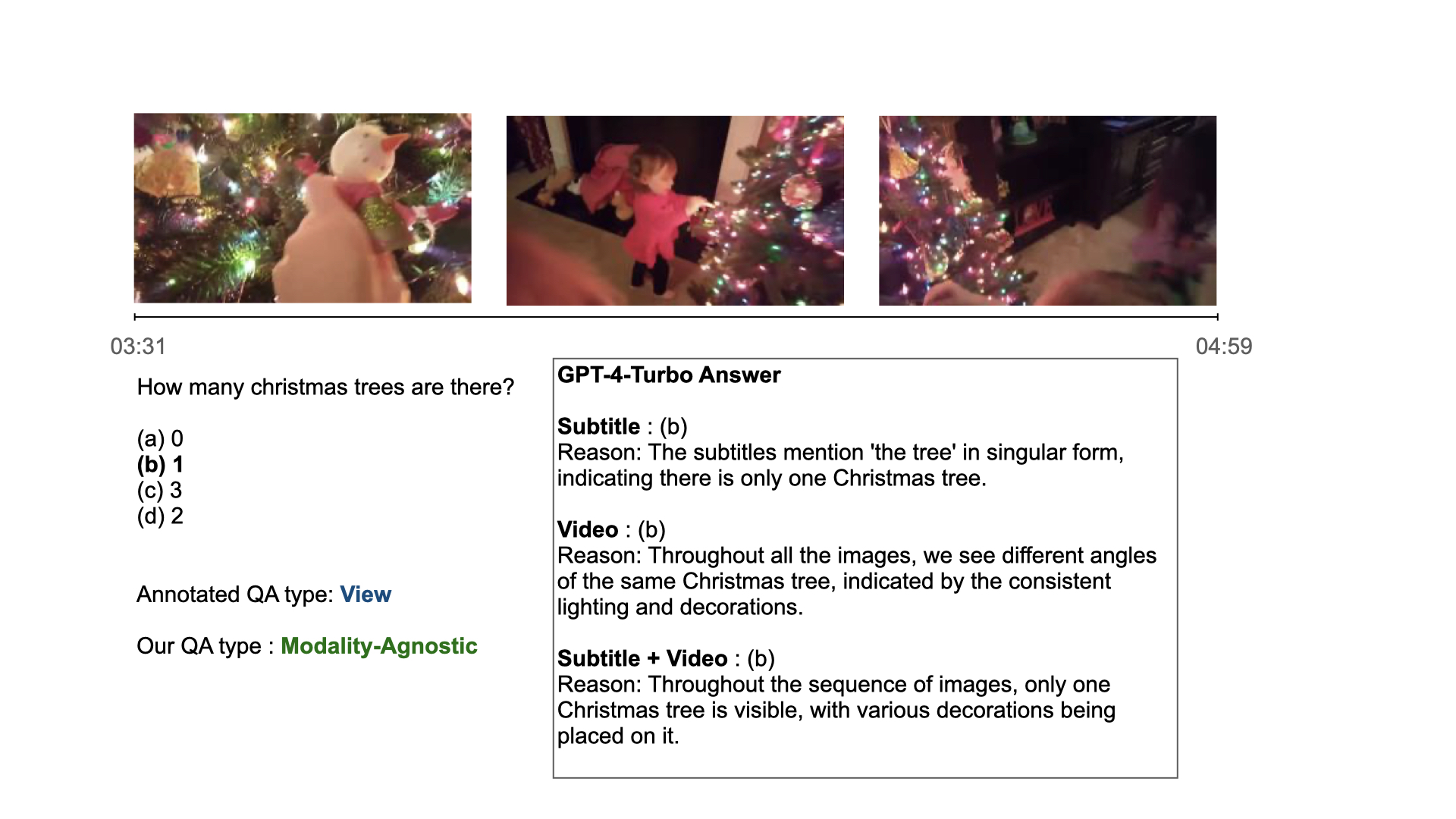}
        \caption{Annotated as ``View''. Subtitle mentions ``[mom] yeah are you gonna put  her this side of the tree'' and ``[mom] haha oh oh Lia you can’t [inaudible] on the tree''.}
        \label{fig:lifeqa_v_ma}
    \end{subfigure}
    \caption{LifeQA: Modality-Agnostic Correct questions that were annotated as (a) ``Both'', (a) ``Sound'', (a) ``View''}
    \label{fig:lifeqa_ma}
\end{figure}

\begin{figure}[!htb]
    \centering
    \begin{subfigure}[b]{0.6\textwidth}
        \centering
        \includegraphics[width=\textwidth]{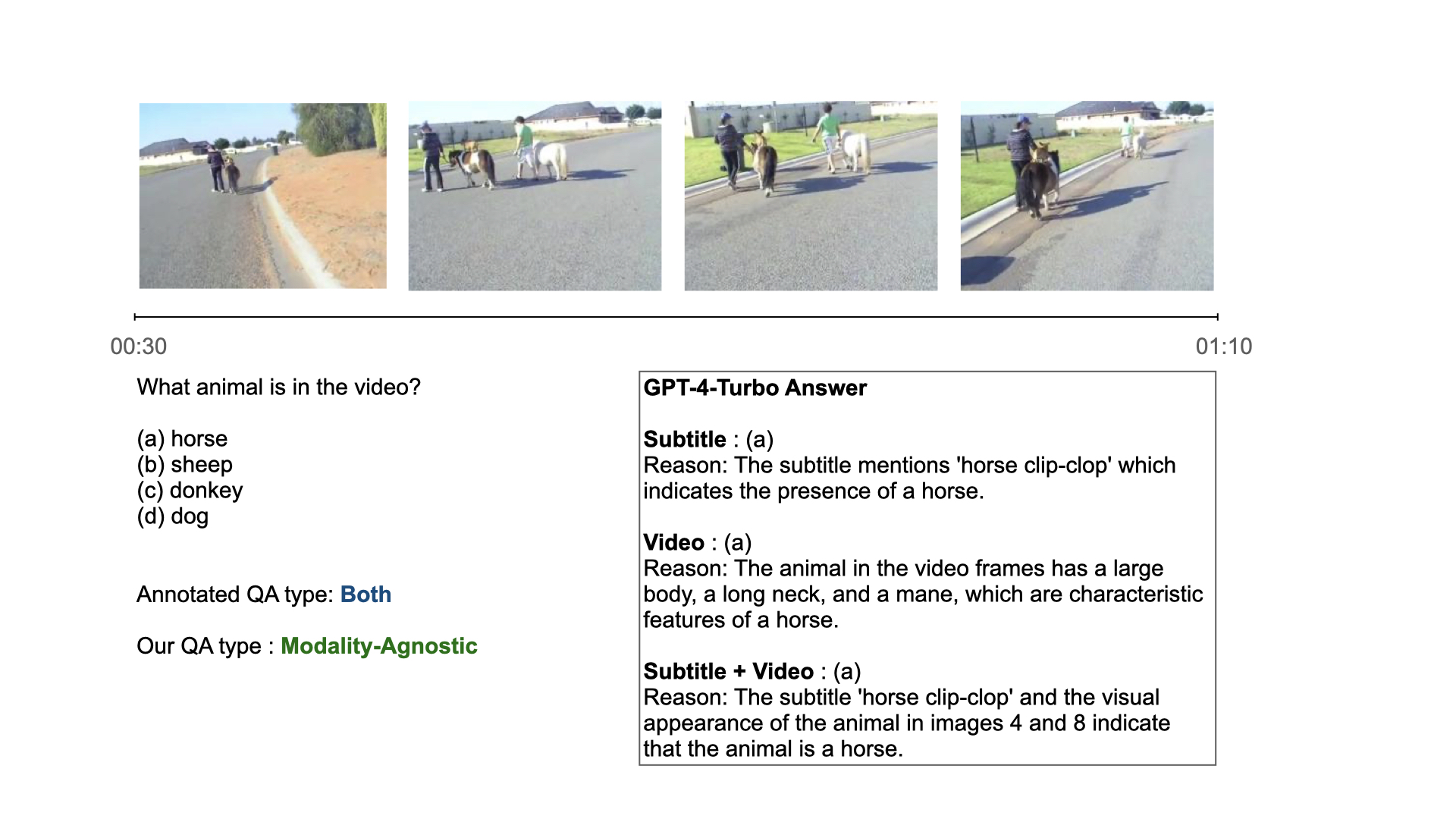}
        \caption{Annotated as ``Both''. Subtitle mentions ``horse clip-clop''.}
        \label{fig:avqa1}
    \end{subfigure}
    \hfill
    \begin{subfigure}[b]{0.6\textwidth}
        \centering
        \includegraphics[width=\textwidth]{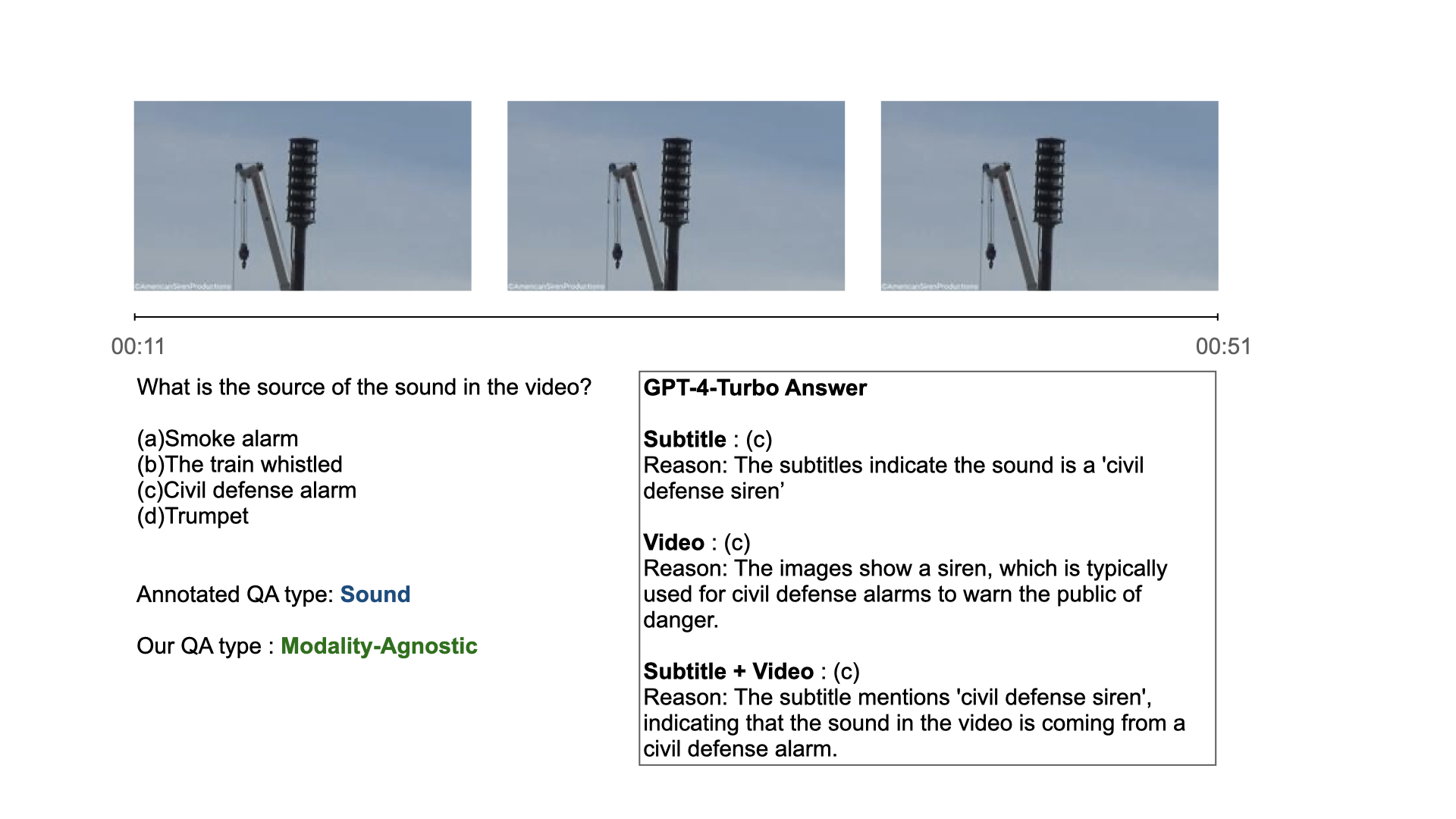}
        \caption{Annotated as ``Sound''. Subtitle mentions ``civil defense alarm''.}
        \label{fig:avqa3}
    \end{subfigure}
    \hfill
    \begin{subfigure}[b]{0.6\textwidth}
        \centering
        \includegraphics[width=\textwidth]{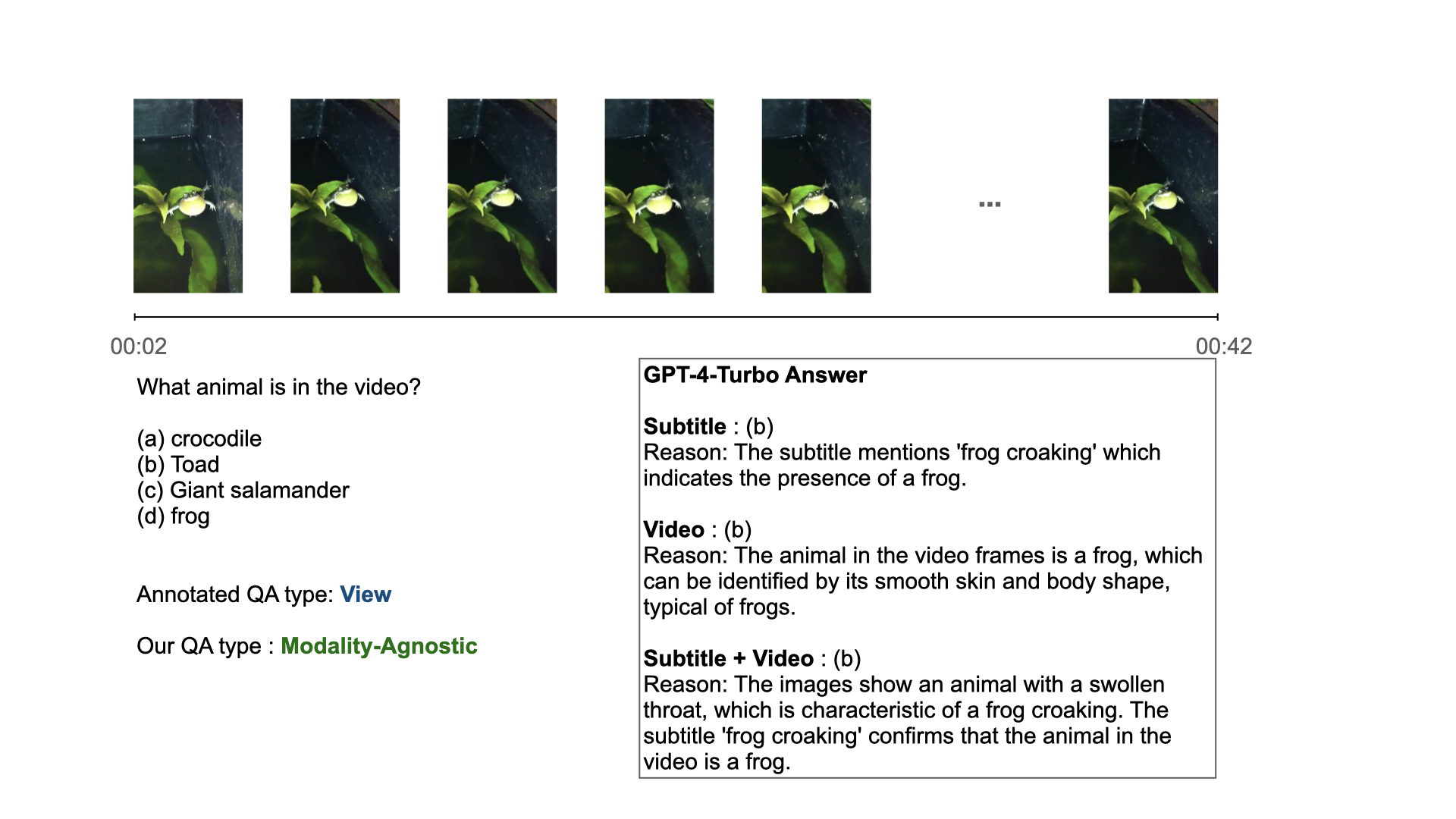}
        \caption{Annotated as ``View''. Subtitle mentions ``frog croaking''.}
        \label{fig:avqa4}
    \end{subfigure}
    \caption{AVQA: Modality-Agnostic Correct questions that were annotated as (a) ``Both'', (a) ``Sound'', (a) ``View''}
    \label{fig:avqa_ma}
\end{figure}

\begin{figure*}[!htb]
    \centering
    \begin{subfigure}[b]{0.6\textwidth}
        \centering
        \includegraphics[width=\textwidth]{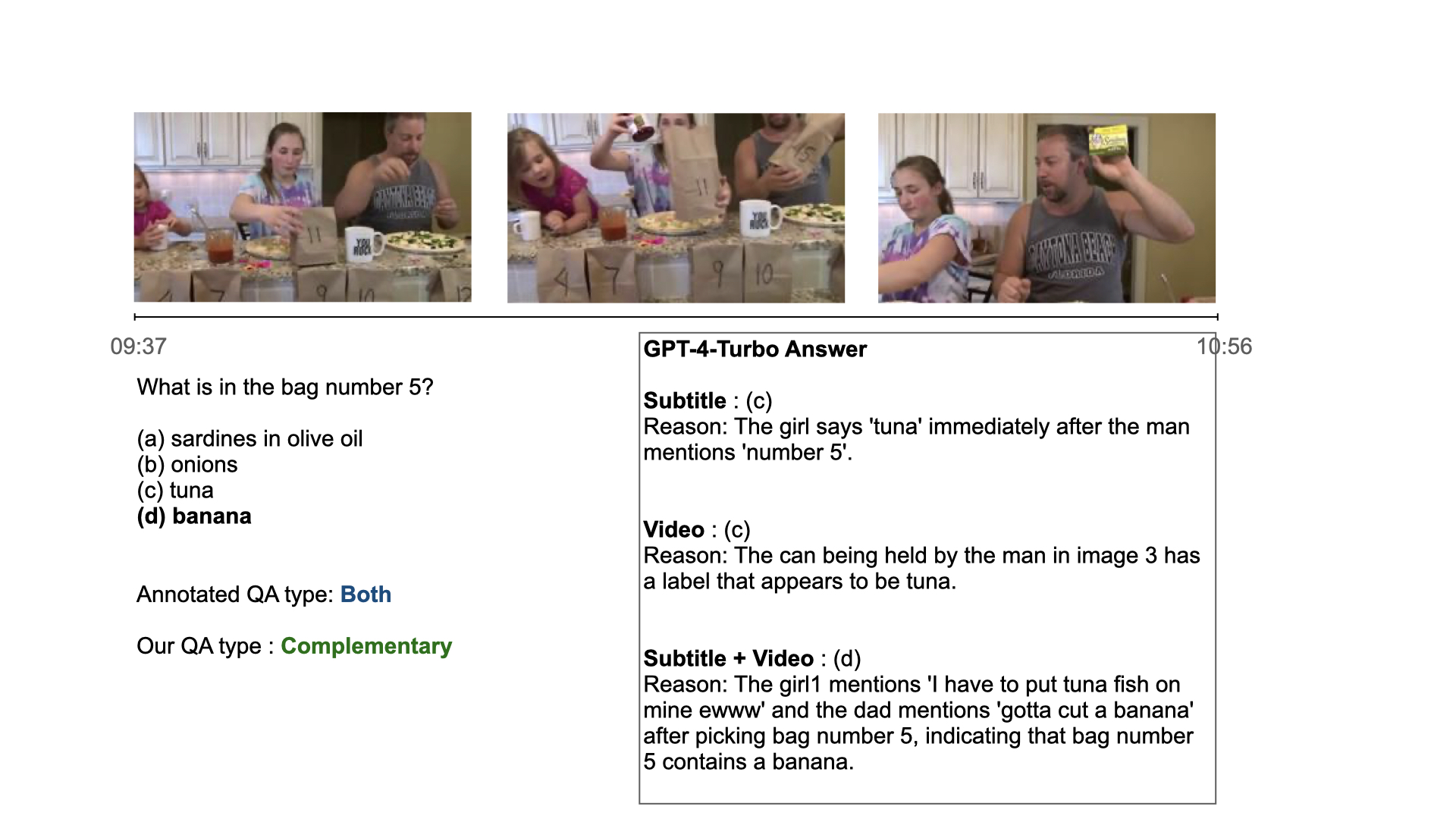}
        \caption{}
        \label{fig:lifeqa_b_c}
    \end{subfigure}
    \hfill
    \begin{subfigure}[b]{0.38\textwidth}
        \centering
        \includegraphics[width=\textwidth]{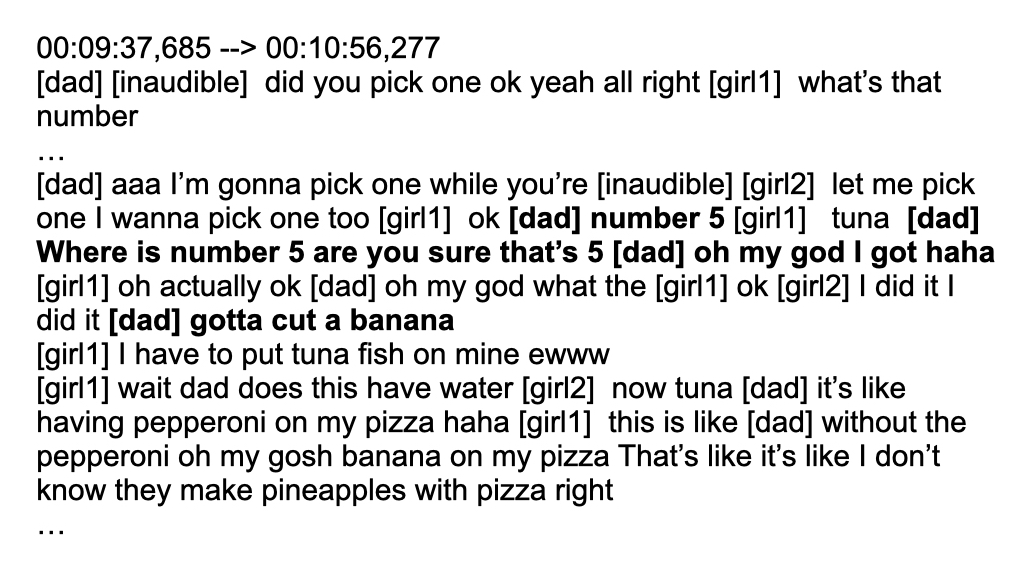}
        \caption{}
        \label{fig:lifeqa_b_c_cap}
    \end{subfigure}
    \caption{LifeQA: Complementary question annotated as ``Both'. Video frames and GPT-4's Answer shown in~\ref{fig:lifeqa_b_c} (left) and subtitles in~\ref{fig:lifeqa_b_c_cap} (right)}
    \label{fig:lifeqa_c}
\end{figure*}
\begin{figure}[!htb]
    \centering
    \includegraphics[width=0.6\textwidth]{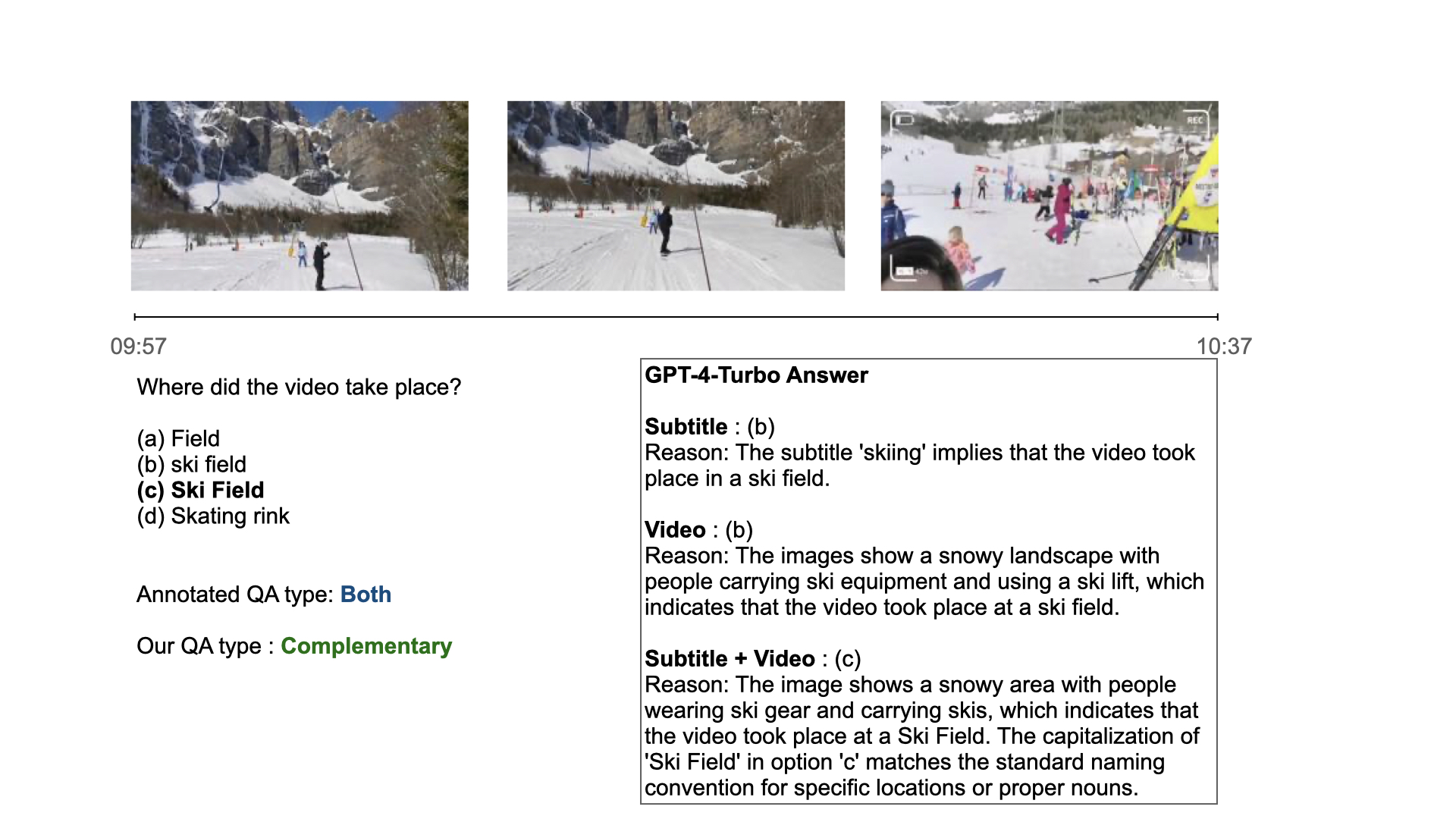}
    \caption{AVQA: Complementary question annotated as ``Both''. Subtitle mentions ``skiing''}
    \label{fig:avqa_c}
\end{figure}

In LifeQA dataset, as shown in Figure~\ref{fig:lifeqa_ma}, we observed various scenarios where GPT-4 provided modality-agnostic correct responses. These include cases where direct answers were present in both modalities, as well as where one modality offered a direct answer (Figure~\ref{fig:lifeqa_b_ma} ) while the other allowed for indirect inference (Figure~\ref{fig:lifeqa_s_ma} and ~\ref{fig:lifeqa_v_ma}). In AVQA dataset, as illustrated in Figure~\ref{fig:avqa_ma}, exhibited a slightly different pattern in its modality-agnostic correct examples. We found that object sound labels provided as subtitles in these videos typically aligned well with image content, thus presenting strong signals from both modalities.The modality-agnostic questions in both datasets highlights the redundancy of information across modalities, suggesting that current datasets may not be optimally designed to challenge models' multimodal integration capabilities.

 We also examine complementary examples from both datasets. While we anticipated that complementary questions would require combining weak signals from both modalities to be answerable, our findings revealed a more complex cases. In Figure ~\ref{fig:lifeqa_c} from LifeQA, we discovered that the video had incorrect start and end timestamps in the annotation. Although the manual captions provided by the dataset appear to be extracted from the correct video segment, the actual video frames showed significant misalignment and therefore did not contain the relevant information. Interestingly, despite these misaligned modalities, we observed that having both information sources actually aided GPT-4 in focusing on the broader context within the subtitle, allowing it to infer details about character actions that weren't explicitly stated. Conversely, the complementary example from AVQA in Figure~\ref{fig:avqa_c} was simply the result of a random selection between two correct choices. 
 
 The scarcity of complementary questions in both datasets limited our ability to analyze how models integrate information from multiple modalities. However, the LifeQA example demonstrates that combining weak signals from misaligned modalities can lead to correct answers, suggesting the potential of complementary questions in fostering effective multimodal integration. This highlights the need for more complementary questions in multimodal datasets, which could push the boundaries of model capabilities in integrating diverse information sources and drive advancements in multimodal reasoning.

\subsection{Configuration for Evaluated Multimodal Models}\label{subsection:appendix_configuration_mm}
Below are the specific models and configurations used in our evaluation of four models:

\paragraph{Merlot Reserve Model~\cite{zellers2022merlot}}
We used the base model fine-tuned on TVQA. Our configuration followed the original Merlot Reserve implementation: we extracted 8 frames and the corresponding subtitle from a 35-second window centered around the middle of the timestamp.

\paragraph{FrozenBiLM~\cite{yang2022frozenbilm}}
We selected a model pretrained with a frozen DeBERTa-V2-XLarge language model and fine-tuned on the TVQA dataset. Following the original experimental set up, we use 10 frames for every clip and subtitles from localized timestamp.  

\paragraph{Llama-VQA~\cite{ko-etal-2023-large}}  For our evaluation, we used the Llama 7B as the base model, with a checkpoint fine-tuned on TVQA. For each video clip, we processed 10 frames as input to the model and subtitles from localized timestamp. As other models, we followed the original implementation.

\paragraph{MiniGpt4-Video~\cite{ataallah2024minigpt4}} 
We evaluated the Llama2-7B based version of MiniGPT4-Video, which processes 45 frames per video and subtitles from localized timestamp. Note that this model was not fine-tuned on the TVQA dataset, unlike the other models in our evaluation. For assessment, we used the evaluation script provided by MiniGPT4-Video, which utilizes GPT-3.5 to compare the predicted answer with the ground-truth.

\subsection{Modality Dependence Analysis through Complementary and Modality Agnostic Questions}\label{subsection:MAC_C_analysis}

\begin{table*}[!htbp]
\centering
\begin{tabular}{l|ccc|ccc}
\toprule
\multirow{2}{*}{} & \multicolumn{3}{c|}{Modality-Agnostic} & \multicolumn{3}{c}{Complementary}\\ 
        & Orig.  & SP ($\Delta$) & VP ($\Delta$) & Orig. & SP ($\Delta$) & VP ($\Delta$) \\ \midrule
Merlot R\text{*} & 90.8 $\pm$ 0.0 & 54.6 $\pm$ 0.8 (-36.2) & 80.0 $\pm$ 0.3 (-10.8) & 18.0 $\pm$ 0.0 & 19.3 $\pm$ 0.7 (+1.3) & 19.7 $\pm$ 1.1 (+1.7)\\ \hline
FrozenBiLM & 94.7 $\pm$ 0.0 & 54.6 $\pm$ 1.4 (-40.1) & 88.8 $\pm$ 0.3 (-5.9) & 45.0 $\pm$ 0.0 & 32.0 $\pm$ 6.3 (-13.0) &  40.7 $\pm$ 2.7 (-4.3)\\ \hline
Llama-VQA & 89.3 $\pm$ 0.0 & 60.0 $\pm$ 21.7 (-29.3) & 87.1 $\pm$ 10.2 (-2.1) & 46.7 $\pm$ 0.0 & 37.5 $\pm$ 2.8 (-9.1) & 47.0 $\pm$ 1.0 (+0.3)\\ \hline
MiniGPT4\text{*} & 65.0 $\pm$ 0.4 & 53.4 $\pm$ 2.8 (-20.5) & 62.3 $\pm$ 1.8 (-2.7) & 49.1 $\pm$ 0.6 & 42.4 $\pm$ 1.6 (-6.7) & 44.2 $\pm$ 3.0 (-4.9)\\ \midrule
Average & 84.9 $\pm$ 0.1 & 53.4 $\pm$ 6.7 (-31.5) & 79.6  $\pm$ 3.2 (-5.4) & 39.7 $\pm$ 0.1 & 32.8 $\pm$ 2.9 (-6.9) & 37.9 $\pm$ 1.9 (-1.8)\\  \bottomrule 
\end{tabular} 
\caption{Accuracy (\%) comparison after feature permutation with five random seeds (Orig: Original, SP: Subtitle permuted, VP: Video permuted, Merlot R\text{*}: Merlot Reserve, MiniGPT4\text{*} : MiniGPT4-Video). All models except for MiniGPT4-Video were fine-tuned on TVQA dataset. }
\label{tab:Mm_permute_evaluation_mac_c}
\end{table*}

To quantitatively assess models' modality integration capabilities, we conducted two sets of experiments focusing on modality-agnostic and complementary questions. As shown in Table~\ref{tab:Mm_permute_evaluation_mac_c}, while models achieve high accuracy on modality-agnostic questions with access to both modalities, their performance exhibits asymmetric degradation under different permutation conditions. Specifically, accuracy drops significantly under subtitle permutation, while remaining relatively robust under video permutation. This contrast in performance degradation indicates that models predominantly rely on textual information, even for questions where both modalities contain relevant signals.

To further investigate models' true multimodal reasoning capabilities, we generated and validated 300 complementary questions from TVQA that explicitly require information from both modalities for correct answering. These questions were generated using Claude and verified with GPT4-Turbo to ensure their complementary nature. The evaluation results in Table~\ref{tab:Mm_permute_evaluation_mac_c} revealed significant limitations in current models' multimodal integration abilities: Merlot Reserve achieved only 18\% accuracy, while other models performed moderately better but still struggled, with accuracies less than 50\%. These notably low performance metrics on complementary questions, coupled with the observation from modality-agnostic questions, show that current models struggle with true multimodal reasoning tasks that require balanced integration of information across modalities.

\subsection{Cross-MLLM Validation of Question Type Classification}\label{subsection:cross_MLLM_validation}

\begin{table}[h]
\centering
\begin{tabular}{c|c|ccc|ccc|ccc}
\toprule
\multirow{2}{*}{\begin{tabular}[c]{@{}c@{}}OL\end{tabular}} & \multirow{2}{*}{Model} & \multicolumn{3}{c|}{TVQA} & \multicolumn{3}{c|}{LifeQA} & \multicolumn{3}{c}{AVQA} \\
\cline{3-11}
 &  & SB (\%) & VB (\%) & MA\_C (\%) & SB (\%) & VB (\%) & MA\_C (\%) & SB (\%) & VB (\%) & MA\_C (\%) \\
\midrule
\multirow{2}{*}{SB} & Claude & 75 & 0 & 22 & 63 & 0 & 33 & 50 & 0 & 50 \\
 & GPT-4o & 51 & 1 & 45 & 55 & 1 & 41 & 42 & 0 & 58 \\
\hline
\multirow{2}{*}{VB} & Claude & 5 & 46 & 34 & 0 & 83 & 5 & 0 & 89 & 11 \\
 & GPT-4o & 0 & 69 & 29 & 0 & 94 & 5 & 0 & 95 & 5 \\
\hline
\multirow{2}{*}{MA\_C} & Claude & 38 & 4 & 54 & 13 & 2 & 82 & 4 & 2 & 94 \\
 & GPT-4o & 8 & 4 & 85 & 11 & 7 & 81 & 2 & 2 & 96 \\
\bottomrule
\end{tabular}
\caption{Comparison of model performance (accuracy \%) across different datasets (OL: Original Label (GPT4-Turbo))}
\label{tab:mllms_compare}
\end{table}

To validate our MIS-based question classification methodology, we conducted additional experiments using multiple MLLMs including Claude-Sonnet~\cite{claude} and GPT-4o, evaluating approximately 300 questions from each dataset. As shown in Table 6, which compares question categorizations using different models against GPT-4 Turbo's original classifications, we observed consistent patterns despite some model-specific variations. Our analysis revealed minimal disagreement between models for opposing unimodal biases (Subtitle-biased vs Video-biased), suggesting robust identification of clearly modality-dependent questions.

When examining cases where models differed, we found that questions classified as unimodal-biased by GPT-4 Turbo were sometimes labeled as Modality-Agnostic Correct (MA\_C) by other models. These differences typically stemmed from models' varying capabilities in language or visual understanding for specific question types rather than fundamental disagreements in modality importance assessment. While GPT-4o demonstrated stronger overall capabilities in both video and language understanding, leading to identification of more MA\_C questions, we cannot definitively assert superior performance for any single model as each succeeded on different subsets of questions.

Importantly, despite these model-specific variations, our key finding remains consistent across all MLLMs: current datasets contain a significant proportion of modality-agnostic questions while truly complementary questions remain scarce. This cross-model validation strengthens our confidence in the broader conclusions about dataset composition and modality bias.

\subsection{Answer Choice Order Bias}\label{MAC_C_analysis}\label{subsection:order_bias}
While our study focuses on dataset-level modality bias, it is crucial to verify that our MLLM-based evaluation methodology is not influenced by model-level biases, particularly potential memorization of answer choice distributions or data leakage~\cite{}. Such biases could compromise our ability to accurately measure the inherent modality biases in the original benchmarks.

To validate against potential answer option order bias, we conducted an experiment using 109 randomly selected questions from TVQA. We evaluated these questions with different permutations of their answer choices to assess whether the order affected the MLLM's modality importance assessment. The results showed only minor variations in question categorization across different answer choice orderings: subtitle-biased questions varied from 37 to 42, video-biased from 33 to 34, with other categories showing minimal changes of 1-2 questions. These small fluctuations were comparable to the natural variance that occurs due to MLLMs' non-deterministic response generation when running our experiments multiple times with identical parameters, indicating that the model's modality importance assessments are not significantly influenced by memorized answer choice orderings.